\documentclass{article}

\usepackage{PRIMEarxiv}

\usepackage[utf8]{inputenc} 
\usepackage[T1]{fontenc}    
\usepackage{hyperref}       
\usepackage{url}            
\usepackage{booktabs}       
\usepackage{amsfonts}       
\usepackage{nicefrac}       
\usepackage{microtype}      
\usepackage{lipsum}
\usepackage{fancyhdr}       
\usepackage{graphicx}       

\usepackage{tikz} 
\usepackage{xcolor} 
\usepackage[breakable]{tcolorbox}
\usepackage{setspace}
\usepackage{multicol}
\usepackage{wrapfig}
\usepackage{graphicx}
\usepackage{subcaption}
\usepackage{authblk}
\usepackage{amsmath}

\graphicspath{{media/}}     

\title{SwarmBrain: Embodied agent for Real-time Strategy game StarCraft II via large language models}

\author[1]{Xiao Shao \thanks{ramsayxiaoshao@gmail.com, Ramsay.Shao@partner.bmwgroup.com}}
\author[1]{Weifu Jiang \thanks{Violet.Jiang@partner.bmwgroup.com}}
\author[1]{Fei Zuo \thanks{Eric.Zuo@partner.bmwgroup.com}}
\author{Mengqing Liu \thanks{pieckliu@gmail.com, mengqingliu365@163.com}}
\affil[1]{AI Team, BMW Archermind Technology Company (BATW, BMW Group)}

\date{} 

\begin{document}
\maketitle

\onehalfspacing

\begin{abstract}
Large language models (LLMs) have recently garnered significant accomplishments in various exploratory tasks, even surpassing the performance of traditional reinforcement learning-based methods that have historically dominated the agent-based field. 
The purpose of this paper is to investigate the efficacy of LLMs in executing real-time strategy war tasks within the StarCraft II gaming environment. 
In this paper, we introduce SwarmBrain, an embodied agent leveraging LLM for real-time strategy implementation in the StarCraft II game environment.
The SwarmBrain comprises two key components: 1) a Overmind Intelligence Matrix, powered by state-of-the-art LLMs, is designed to orchestrate macro-level strategies from a high-level perspective. This matrix emulates the overarching consciousness of the Zerg intelligence brain, synthesizing strategic foresight with the aim of allocating resources, directing expansion, and coordinating multi-pronged assaults.
2) a Swarm ReflexNet, which is agile counterpart to the calculated deliberation of the Overmind Intelligence Matrix. Due to the inherent latency in LLM reasoning, the Swarm ReflexNet employs a condition-response state machine framework, enabling expedited tactical responses for fundamental Zerg unit maneuvers. 
In the experimental setup, SwarmBrain is in control of the Zerg race in confrontation with an Computer-controlled Terran adversary. 
Experimental results show the capacity of SwarmBrain to conduct economic augmentation, territorial expansion, and tactical formulation, and it shows the SwarmBrain is capable of achieving victory against Computer players set at different difficulty levels.
Specifically, SwarmBrain's success rate is 100\% against Computer adversaries at the Very Easy, Easy, Medium, and Medium Hard levels. Furthermore, even at the Hard level, SwarmBrain sustains a substantial winning percentage, securing victories in 76\% of the matches.

\end{abstract}

\keywords{Large Language Model \and Real-Time Strategy \and StarCraft II \and Embodied Agent \and SwarmBrain \and Overmind Intelligence Matrix \and Swarm ReflexNet }

\section{Introduction}
"\textit{I have no name, but you may address me as Swarm. I am one of its castes. My specialty is intelligence. You sought to breed us. Use us. But your crude experiments triggered certain genetic protocols, and I was born. I am only a few weeks old, but I have millions of years of racial memory.
I’m just a tool. One the Swarm has used to deal with threats like yours many times. Through her memories, I understand yours race. An especially vigorous one. I expect they could be here, competing with us within a few hundred years. But in the timescale the Swarm operates, your race will soon be gone. Most likely, you’ll destroy yourselves. Intelligence is not a winning survival trait.}"

\hfill --- Love, Death \& Robots (Season 3 Episode 6 “Swarm”)

StarCraft II \footnote[1]{https://starcraft2.blizzard.com/}, launched by Blizzard Entertainment in 2010, is a real-time strategy (RTS) game that has garnered substantial attention within the gaming community. Participants in standard gameplay competitions have the opportunity to engage in strategic contests while playing the roles of one of three distinct races: Terran, Zerg, and Protoss. The unique gameplay mechanics and complex strategic depth of StarCraft II have established it as a robust experimental platform for the progression of artificial intelligence (AI), making it a subject of considerable interest in technology and AI research fields (\cite{[ai1]}-\cite{[ai7]}).

Reinforcement learning (RL) \cite{[1]}-\cite{[10]} is the most popular method for training AI agents to make a sequence of decisions by interacting with a complex environment to achieve a specific set goal. By receiving the feedback of rewards or penalties, RL-based agents are capable of learning from the experiences, and optimizing their behavior to maximize the cumulative rewards over time. DeepMind's AlphaGo \cite{[11]} has marked a significant milestone in the field of RL with its groundbreaking achievements. Then, AlphaStar \cite{[12]} from DeepMind is another testament to the prowess of RL in mastering StarCraft II with complex environment, and defeated many professional players.

Despite the traditional RL-based agents \cite{[12]}-\cite{[rla4]} achieved significant performance in StarCraft II, it still encounters considerable challenges when tasked with achieving high-level proficiency in such complex environments. The primary complication arises from the attempt to directly map extended, complex objectives to the lowest-level actions of keyboard and mouse inputs. This low-level direct mapping strategy frequently falls short of capturing the comprehensive dynamics of the battlefield. In contrast, large language models (LLM) \cite{[llm1]}-\cite{[llm5]}, by virtue of their inherent capacity for high-level abstraction coupled with their facility in understanding intricate contexts, can afford a superior macroscopic comprehension of the entire battlefield situation. Such a holistic perspective enables the AI agents \cite{[npc]}-\cite{[aia4]} to devise tactical decisions that are more coherent and informed, potentially improving their performance and adaptability within intricate scenarios.

However, employing LLM directly in the context of RTS games such as StarCraft II presents a significant challenge due to the game's inherent demand for prompt decision-making. Previous implementations of LLM-based agents \cite{[llma1]},\cite{[llma2]} have achieved notable breakthroughs in exploration-oriented tasks, notably in environments like Minecraft. These achievements are primarily attributed to the relatively relaxed real-time constraints of such tasks. However, the landscape is drastically different in StarCraft II, where the ability to react swiftly under varying scenario conditions is critical. Average players typically maintain an Actions-Per-Minute (APM) rate around 100, while more advanced players reach upwards of 200 APM. During intense gameplay, it's not uncommon for a player's APM to surge to 300-400, equating to a formidable 5 to 6 commands executed every second. Expecting the LLM to match this operational tempo is currently unrealistic. For instance, ChatGPT 4.0 \cite{[llm3]} might take upwards of 20 seconds to process a single response containing 2000 tokens, a duration far too lengthy for the rapidly shifting dynamics of the StarCraft II battlefield. In summary, the latency inherent in the state-of-the-art LLM processing precludes their direct application in highly time-sensitive environments like competitive StarCraft II, necessitating novel approaches for adapting such models to keep pace with the game's exigencies.

Given the shortcomings of LLMs in RTS environment as seen in StarCraft II, this leads us to consider the social structure of the Swarm, the integrity of order is conserved without necessitating high levels of individual cognition among its constituents. Each unit of the Swarm is imbued with a set of predetermined functions, mirroring the organizational paradigm of the Swarm individuals. In times when the Swarm species faces external threats, the situation calls for the engagement of a high-level Swarm intelligence that conceptualizes retaliatory strategies from a macroscopic viewpoint, much in the way a human military commander devises plans. Following such strategic development, units within the Swarm implement the distributed tasks in accordance with directives from the collective intelligence, thereby mounting a defense against intruders.

To mimic this process, the SwarmBrain for mastering the RTS game StarCraft II as the Zerg race is introduced. SwarmBrain consists of two key components: 1) a \textbf{Overmind Intelligence Matrix} that is designed to orchestrate macro-level strategies from a high-level perspective. 2) a \textbf{Swarm ReflexNet} that aims to imitate the intelligence intrinsic of Zerg individuals for fundamental Zerg unit maneuvers. 

To be specially, the Overmind Intelligence Matrix is designed to formulating macro strategies based on a comprehensive understanding of the battlefield dynamics. It consists of two parts: a Overmind Brain and a SC2 Brain, both powered by LLMs. The Overmind Brain mimic the intrinsic consciousness of the Swarm intelligence brain, takes into account the state of the agent, the adversary's status, and the comprehensive battlefield intelligence, synthesizing strategic foresight with the aim of allocating resources, directing expansion, and orchestrate offensive engagements against adversaries. Since LLM struggle to correctly handle all the task in one shot, the SC2 Brain is employed to translate natural language-based tactical concepts from Overmind Brain into actionable commands with the StarCraft II. 
Due to the slow reasoning speed of LLM, which hinders effective engagement within the fast-paced RTS game environments, and their inability to issue highly detailed operational commands due to the lack of visual information input, an LLM-based agent approach faces significant challenges. To address these issues, the Swarm ReflexNet is introduced, which endows individual Zerg units with simple, autonomously executed tasks. These tasks include prioritizing attack targets, reactive protocols when under assault, and the Queen's consistent larva-spawning behavior, etc.

\section{Related Work}
\label{sec:headings}

\subsection{Large language models}
With the emergence of ChatGPT \cite{[llm3]}\cite{[chatgpt1]}, the capacity for LLM \cite{[chatgpt2]}-\cite{[glm]} to exhibit remarkable capabilities has been validated, showcasing unique abilities inherent to these expansive models. The mathematical reasoning \cite{[math1]}-\cite{[math3]}, generalization, and adherence to instructions exhibited by LLM have undergone a qualitative enhancement.  Consequently, LLM-based methods are now being employed in more intricate application scenarios. In contrast to proprietary LLMs, some open-source models, such as LLaMA \cite{[llama1]} and LLaMA 2 \cite{[llama2]}, demonstrate formidable emergent capabilities.  Furthermore, small-scale models in the current stage have been proven to possess abilities similar to or even surpassing those language models with large parameters. Specifically, LLMs have demonstrated robust generalization capacities across various specialized domains, such as code generation \cite{[code1]}-\cite{[code5]} and tool usage \cite{[tool1]}-\cite{[tool5]}, exemplified by tools like Toolformer \cite{[tool2]}.

As LLMs gain increasing computational prowess, several prompt techniques have been confirmed to be effective in handling complex tasks, including methods like Chain of Thought (CoT) \cite{[cot]}-\cite{[cot5]} and the ReAct \cite{[42]} approach.  These techniques involve guiding LLMs through in-depth analysis of the input questions before generating output, aiming to maximize the accuracy of results. The advent of LLMs, including ChatGPT and GPT-4, signifies a pivotal step forward in natural language processing \cite{[nlp1]}-\cite{[nlp5]}. These models, characterized by multi-round conversation capabilities, have demonstrated an impressive aptitude for following intricate instructions.  The integration of vision capabilities in GPT-4V \cite{[gpt4v]} further expands the scope of AI applications, enabling tasks ranging from problem-solving and logical reasoning to tool usage, API calls, and coding. Recent studies on GPT-4V highlight its ability to understand various types of images, including simple user interfaces in popular smartphone apps.  However, challenges arise with new apps featuring less typical UIs, underscoring a major problem addressed by ongoing work.  Among open-source endeavors, the LLaMA series stands out, fine-tuned to acquire conversational abilities and employing a decoder-only architecture similar to ChatGPT.  Building upon LLaMA, multimodal LLMs like LLaVA \cite{[llava]}, ChartLlama \cite{[chartllama]}, and StableLLaVA \cite{[stablellava]} also showcase vision understanding capabilities akin to GPT-4V. Despite these advancements, a performance gap persists between open-source models and GPT-4V, suggesting potential areas for further development.

\subsection{Large language models for agent planning}
The advancement of LLMs, particularly in the realm of multimodal language models \cite{[mlm1]}-\cite{[mlm6]}, underscores a remarkable trend towards the development of LLMs as sophisticated self-decision systems. At present, LLM agents have proven effective in various complex downstream tasks.  For instance, for embodied applications like Rt-1 \cite{[rt1]}, Rt-2 \cite{[rt2]}, and Voxposer \cite{[voxposer]}, LLMs serve as decision hubs, propelling robots to accomplish intricate, long-sequence tasks. For exploration-based tasks, such as Minecraft, projects like Voyoger \cite{[voyager]}, Ghost-in-the-minecraft \cite{[ghost]} and Plan4mc \cite{[plan4mc]} utilize agents for self-exploration within sandbox environments, learning valuable skills based on environmental feedback.  Within the realms of mobile assistant applications and web scenarios, agents also exhibit outstanding task generalization capabilities. For instance, AppAgent \cite{[appagent]}, in zero-shot experiments on mobile devices, demonstrates commendable task generalization across ten different apps. In web navigation tasks \cite{[web]}, it executes instructions proficiently. Especially when multimodal models serve as the foundation for agents, their performance in visual and textual tasks becomes more versatile and intelligent, such as Jarvis-1 \cite{[jarvis]}. In general-purpose agent types, AutoGPT \cite{[autogpt]} and HuggingGPT \cite{[hugginggpt]} demonstrate autonomy in diverse tasks.

In the multi-agent domain, agents have been proven to engage in iterative learning in both competitive and cooperative modes.  In social deduction-type agents, generative agents \cite{[npc]}, through interactions with multiple agents, exhibit activities and thinking approximating human-like behavior. In typical software development domains such as code writing and software project management \cite{[software1]}\cite{[software2]}, multi-agent agents showcase the achievement of complex team tasks through collaborative processes.  Additionally, in gaming environments, such as the Werewolf game \cite{[game]}, agents learn human-like skills like disguising and lying.

The use of LLMs as agents for complex tasks has gained attention, exemplified by initiatives like AutoGPT, HuggingGPT, and MetaGPT \cite{[metagpt]}.  These projects showcase capabilities beyond basic language tasks, engaging in activities requiring higher cognitive functions such as software development and gaming.  Innovative approaches, like synergizing reasoning and acting in LLMs, enhance decision-making and interactive capabilities \cite{[decisionmake]}.  Multimodal LLM agents, capable of processing various inputs, further broaden LLM applications, enabling more effective interaction and completion of complex tasks.  The adaptability of agents in various scenarios, including embodied applications, exploration tasks, mobile assistance, web navigation, and gaming, highlights their versatility and intelligence.

\section{SwarmBrain}
\label{sec:others}
The framework of the proposed SwarmBrain is shown in Fig. 1, which consists of: (1) a \textbf{Overmind Intelligence Matrix}, tasked with the formulation of sophisticated, high-level strategic directives, and (2) a \textbf{Swarm ReflexNet}, a subsystem engineered to endow the Zerg's rudimentary units with the capability to execute fundamental operations through conditioned reflexes. 

The interaction between SwarmBrain and StarCraft II environment is exemplified in Fig. 1. Environmental observations are sourced through the python-sc2 \footnote[2]{https://github.com/Dentosal/python-sc2, https://github.com/BurnySc2/python-sc2} API interface, which channels the game state information into both the Overmind Intelligence Matrix and the Swarm ReflexNet. 
Since the obtained game state information contains comprehensive and intricate in-game state information, our methodology involves a selective extraction process whereby only pertinent data are harvested. Subsequently, these extracted data undergo a series of mathematical computations to distill the necessary parameters. The resulting refined information is then encapsulated into natural language, serving as the input for the Overmind Intelligence Matrix.
The Overmind Intelligence Matrix, which is based on the LLM, processes the processed natural language data to formulate strategic directives for the Swarm ReflexNet. Full prompts are presented in Appendix. The Swarm ReflexNet, in turn, utilizes this observation information to execute conditioned reflex-like basic decisions for the Zerg units. The details are introduced in Sec 3.2.

\begin{figure}[htbp]
  \centering
  \label{fig:fig1}
  \includegraphics[width=5.0in]{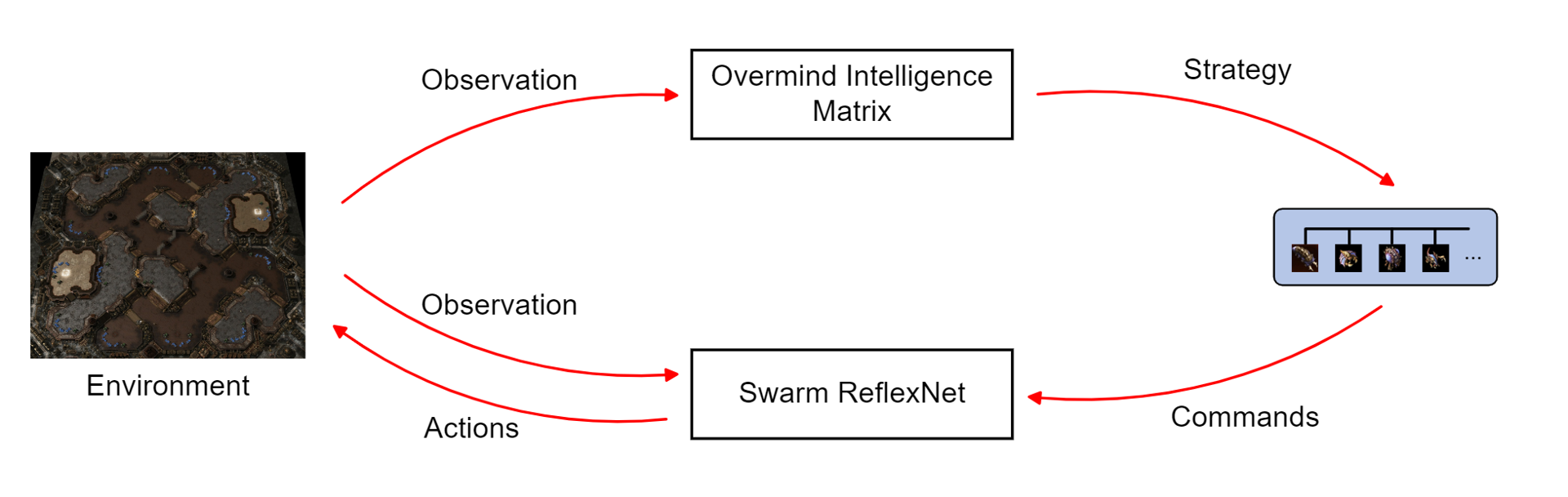}
  \caption{The framework of the interaction between SwarmBrain and StarCraft II environment.}
\end{figure}

\subsection{Overmind Intelligence Matrix}
The overall framework of the Overmind Intelligence Matrix is shown in Fig. 2. It is envisioned as a composite system, comprising four distinct but interrelated components. Each component is designed with specialized functionalities, enabling the Matrix to conduct high-dimensional strategic operations. The components of the Overmind Intelligence Matrix and their respective functionalities are listed as follows:

(1) \textbf{The Overmind Brain}: As the linchpin of the Overmind Intelligence Matrix, this module is designed to emulate the role of the Overmind—the overarching intelligence of the Zerg swarm—tasked with crafting tactical strategies aimed at safeguarding and proliferating the Zerg swarm. 

(2) \textbf{The Text-Based Memory System}: Functioning as the repository of the Overmind Brain's tactical strategies, this subsystem preserves a record of the cognitive processes and tactical strategies previously formulated. The memory system enhances the Overmind Brain's ability to learn from past encounters, refine its further strategic over time, and minimize the duplication of unnecessary instructions.

(3) \textbf{The SC2 Brain}: This crucial interface translates the Overmind Brain's strategic conceptions into executable command sets compatible with the StarCraft II environment. It operates as the translator, transfer the natural language-based strategies into a sequence of tangible, game-specific actions.

(4) \textbf{The Command Center}: It functions as the operational nexus that dispatches the SC2 Brain's command sequences, when the SC2 Brain's command does not meet the execution conditions (for example, the manufacturing conditions are not met), it will be temporarily suspended until the requirements are met before the command is issued.

\begin{figure}[htbp]
  \centering
  \label{fig:fig2}
  \includegraphics[width=6.4in]{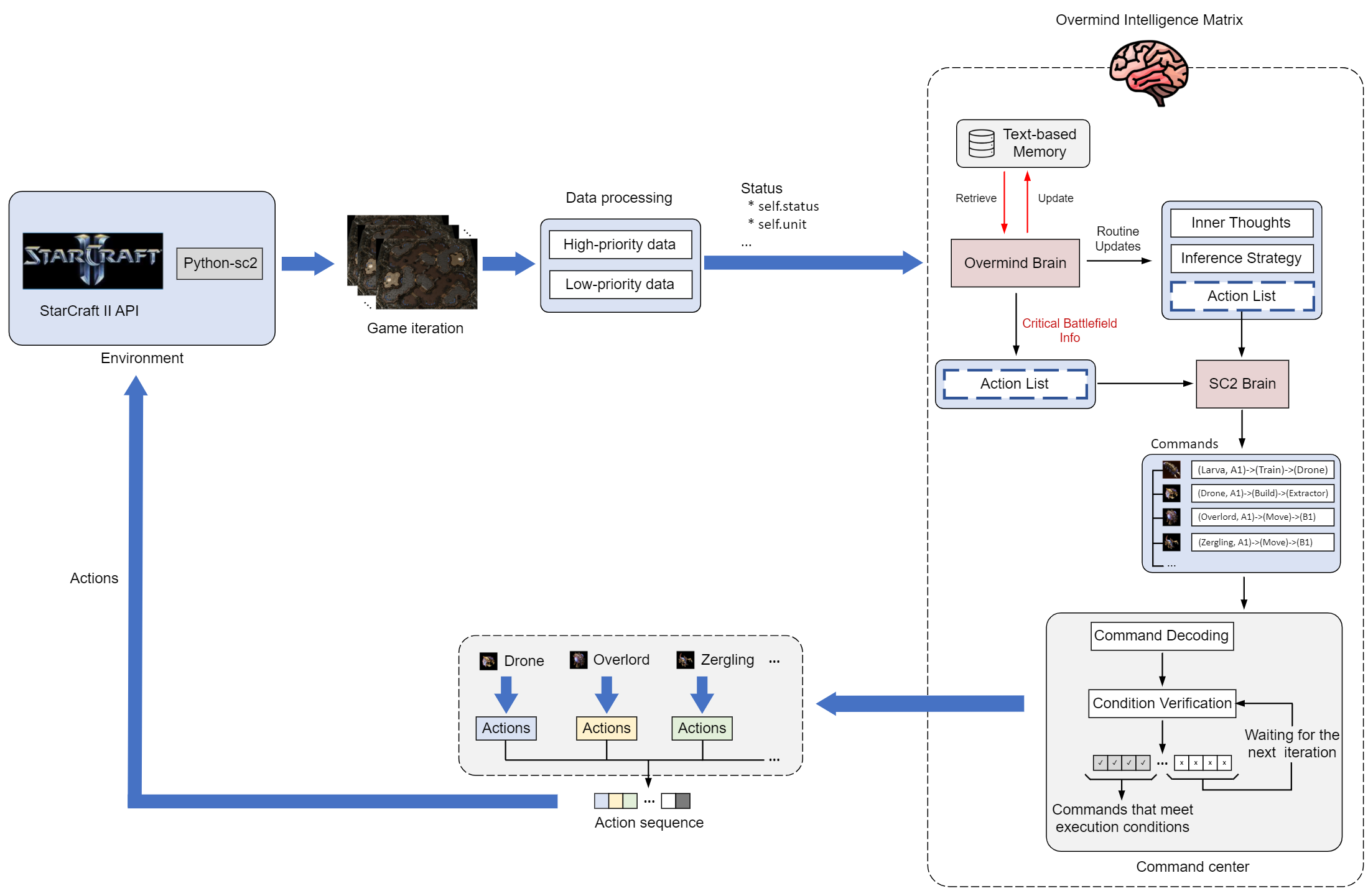}
  \caption{The framework of the Overmind Intelligence Matrix.}
\end{figure}

\subsubsection{The Overmind Brain}
StarCraft II presents a complex and multifaceted environment where standard matches occur within the confines of meticulously crafted square maps. Competitors begin at opposing corners of the map, with players needing to pay close attention to an array of game data. This data encompasses the player's economic status, including mineral reserves and gas supplies, the construction order of structures, the production of units, and enemy's situation. Furthermore, due to the presence of the "Fog of War", it is crucial to constantly monitor the evolving situation on the battlefield. This involves dispatching scouts early on to gather critical intelligence about the opponent's status, thereby allowing predictions of the adversary's tactical strategy based on their construction activities. Such information is vital for the success of human players in a match, yet understanding these data and further, analyzing battlefield conditions to develop coherent and effective tactical strategies, poses significant challenges for LLMs. 

To face these challenges, the Overmind Brain is introduced, which is an innovative concept designed exclusively for the strategic control of the Zerg swarm. By activating the intrinsic "survival instinct" typical of the Zerg, the Overmind Brain is conceived to sustain the integrity of the faction and to respond efficiently to potential external threats. Its purpose does not lie in formulating tactical strategies through the utilization of statistical reasoning inherent in LLMs when presented with an abundance of data, as this is exceedingly challenging. Instead, LLMs excel in role-playing within structured scenarios. By establishing a clear context, the Overmind Brain is proficient at assessing imminent threats and devising appropriate strategies. This narrative sets the stage for introducing a novel agent-based approach in the StarCraft II environment, enhancing the ability to successfully play as the Zerg.
The input information of the Overmind Brain consists of several aspects:

(1) \textbf{Act as an Overmind Brain}. To act as the Overmind Brain, the prompt is set as “\textit{You are an intelligent brain of Zerg swarm in StarCraft II game. You are very aggressive and know all the dependencies between Zerg units, Zerg buildings, and Zerg technological research..}…”.\\
(2) \textbf{Map locational information}. The primary function of map locational information is to convey information about mineral field locations to LLM, thereby facilitating a more comprehensive understanding of the terrain across the entire battlefield. This spatial awareness is essential for LLM to generate strategic insights pertinent to the geographical distribution of resources within the given map.\\
(3) \textbf{The repository of the strategies}, which maintains the histortical tactical strategies, thereby ensuring that new strategies formulated by the LLM align synergistically with past strategy and precluding repetitive commands.\\
(4) \textbf{Comprehensive battle assessment protocols for Overmind Brain}, which aims to enhance the analytical capabilities of the Overmind Brain, enabling it to assess battlefield conditions from a comprehensive perspective. This encompasses evaluations of the current stage of the match, the status of Zerg forces—which includes an inventory of Zerg units and buildings as well as an appraisal of the Zerg technological research. Additionally, an analysis of the prevailing Zerg operational strategies is required. Consideration must also be given to the opponent's situation, covering scrutiny of their units and buildings, their strategic intentions, and potential threats posed to Zerg population. Furthermore, gathering and integration of scouting intelligence are imperative to form a complete battlefield analysis.\\
(5) \textbf{Critical battlefield information}. The purpose of recognizing and prioritizing critical battlefield information is to ensure that the Overmind Brain accords significant attention to pivotal moments when Zerg forces incur engagements with adversary troops. This focus is integral for enabling a swift and effective response to evolving combat scenarios. In situations where the Zerg are subject to strikes from opposing forces, it is paramount that the Overmind Brain processes this crucial battlefield intelligence promptly to facilitate rapid decision-making and adaptation to the tactical landscape. Consequently, enhancing the Overmind Brain's situational analysis capabilities and response mechanisms is essential for maintaining strategic advantage and operational efficacy in the face of hostile engagements.\\
(6) \textbf{The agent’s current situation}. The operational landscape of LLM in RTS environment like StarCraft II is largely defined by the spatial distribution and status of its units and buildings. The effectual command of these assets by LLM necessitates a representation framework that provides concise, state-aware, and contextually relevant information. Current retrieval methods via python-sc2 interface yield raw data that encapsulates unit details inadequately, as typified by entries such as “Unit(name='Overlord', tag=4353163265),” which suffer from verbosity and lack critical state descriptors. Presenting such raw data within an LLM prompt, especially in scenarios entailing large-scale conflicts, introduces several shortcomings: 1) the surplus of superfluous data markedly escalates the LLM's inference latency, undermining RTS analysis. 2) the LLM's inability to discern the state of individual units—specifically whether they are engaged in combat or in idle state—compromises situational awareness and decisional accuracy. 3) the granular management of each unit is unfeasible for LLM devoid of visual inputs. To alleviate the cognitive burden on the LLM and enhance the strategic interaction within the game, we have introduced Swarm ReflexNet, which would be discussed in Sec 3.2.\\
(7) \textbf{Enemy current situation}, including detected enemy units and detected enemy buildings, with the same format in "The agent’s current situation".\\ 
(8) \textbf{Response rules}. Employing the Chain of Thought \cite{[cot]} approach, LLM is guided to perform step-by-step reasoning based on the current battle situation, aligned with the status of both allied and opposing units and buildings, thereby bolstering the accuracy of the inferential process.\\
(9) \textbf{Response format}. The generated list of actions, structured in the JSON format, facilitates the subsequent processing by the command center.\\

\begin{figure}[htbp]
  \centering
  \label{fig:fig3}
  \includegraphics[width=4.5in]{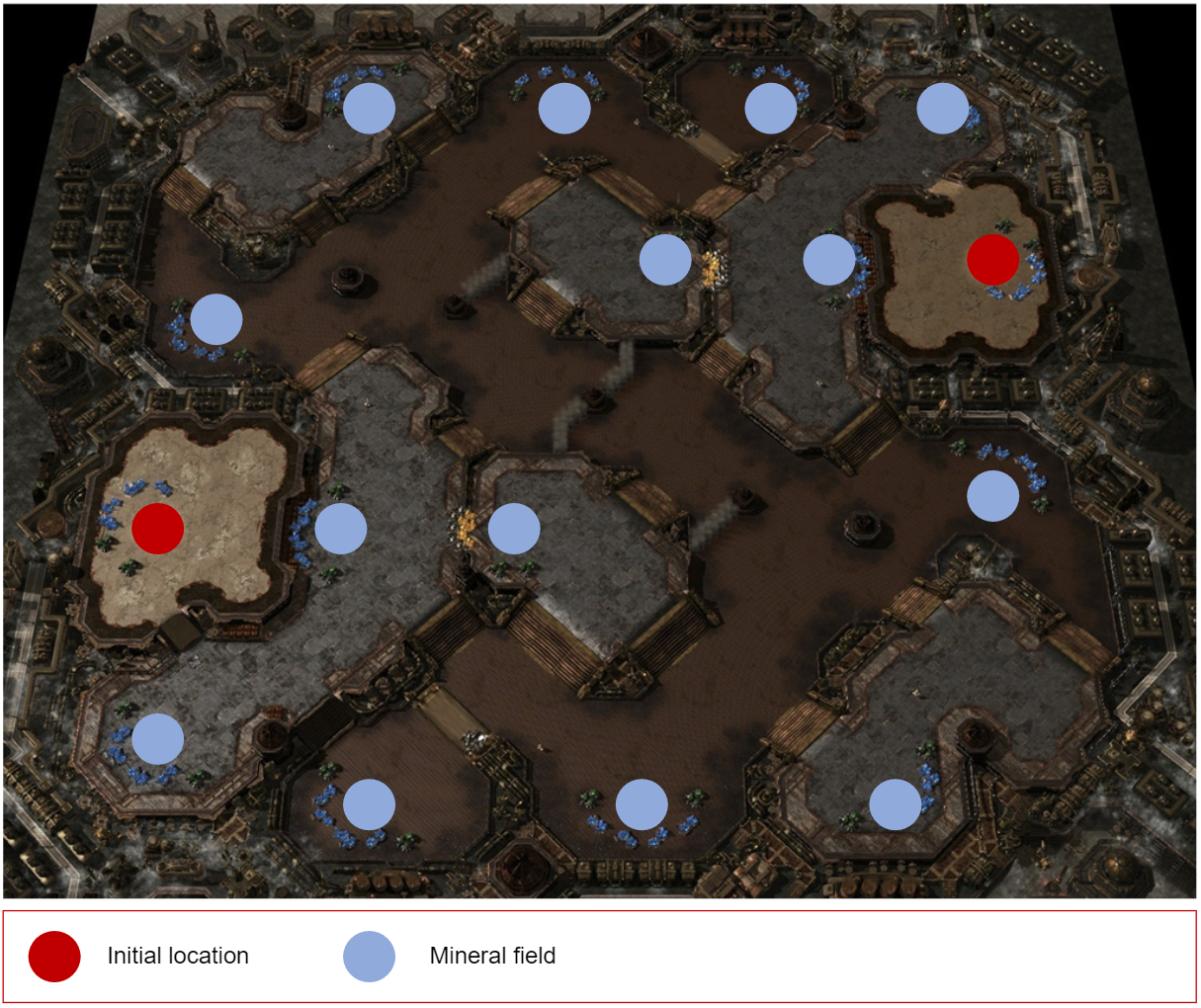}
  \caption{The example map of "Automaton LE".}
\end{figure}

\textbf{Map locational information}

Incorporating the intricate topology of StarCraft II's dynamic battlegrounds is a pivotal aspect of the Overmind Brain's decision-making process. A comprehensive understanding of its spatial positioning relative to the adversary is central to enacting viable strategies. As illustrated in Fig. 3, taking the map "Automaton LE" as an example, the terrain's layout is a rectangular map and the design is complex. To be specially, there are numerous mineral fields embedded within the map, typically allowing for an equitable distribution of eight resource sites per faction under conditions of balanced power.

In a manner akin to strategic board games, the participants—represented by opposing factions—engage in a tactical struggle to seize mineral-rich territories by dispatching military units. The primary objective centers around the eradication of opposing forces through territorial domination, thereby securing control over the map's assets. Such conquests serve not only to diminish the military capacity of the enemy but also to bolster the economic foundation of the conqueror's own faction. Ultimately, by efficiently leveraging these gains to amplify one's economic and military infrastructure, a player can expediently achieve victory.

This elaborate terrain composition presents a significant challenge in terms of conveying the nuanced geographic details to LLM which is at the core of the Overmind Brain. The effectiveness of strategic commands generated by the Overmind Brain is contingent upon its comprehend of the complete map information, which is not readily transmissible through textual descriptions alone. The Overmind Brain, while primarily LLM-based, requires enhanced sensory interpretation capabilities to ascertain a holistic grasp of the game environments—a formidable endeavor, crucial for the issuance of nuanced tactical commands. 

\begin{figure}[htbp]
  \centering
  \label{fig:fig3}
  \includegraphics[width=4.5in]{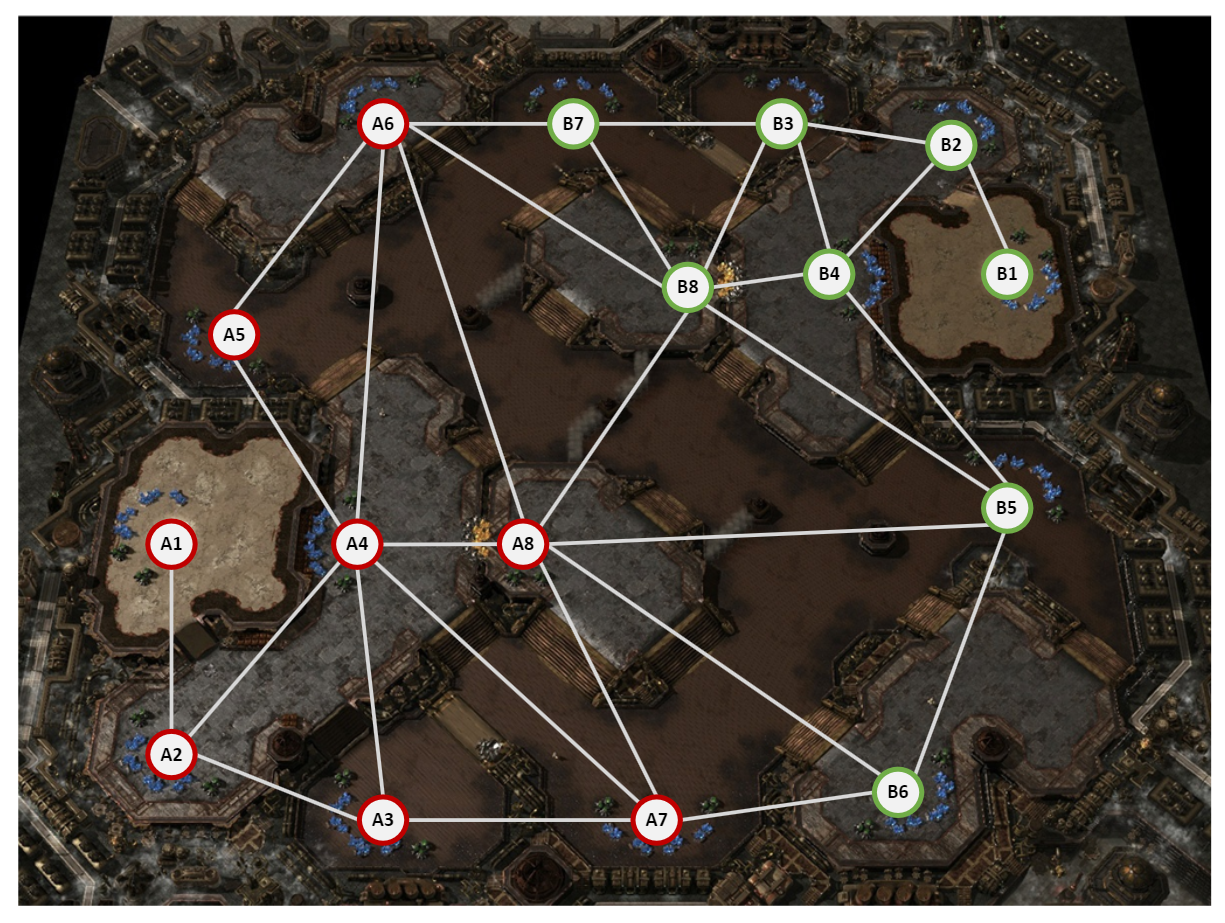}
  \caption{The connection diagram between different mineral fields.}
\end{figure}

\begin{figure}[htbp]
  \centering
  \label{fig:fig3}
  \includegraphics[width=4.5in]{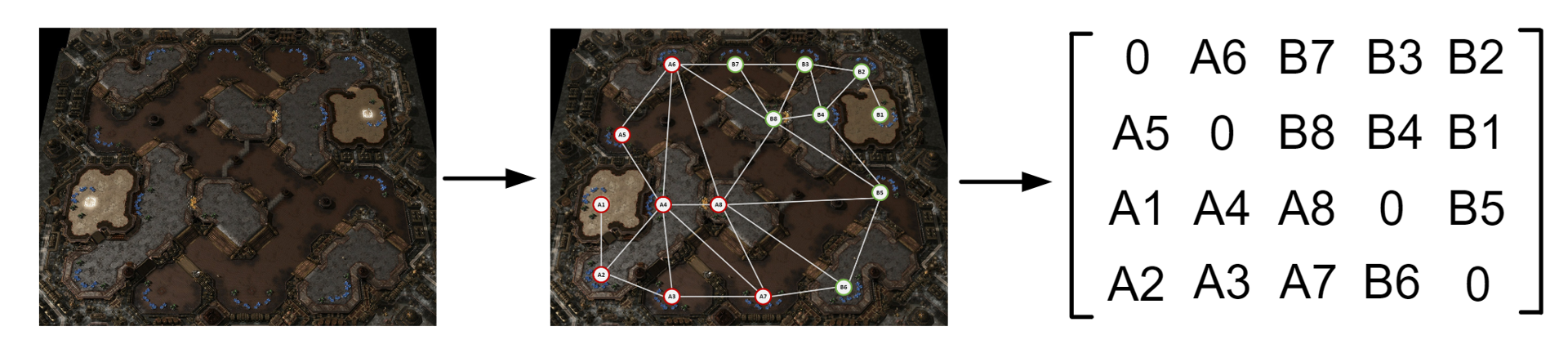}
  \caption{A map matrix representation centered on mineral field locations.}
\end{figure}

To bridge this gap between the spatial complexity of StarCraft II maps and the LLM's understandings, we have utilized a method to transmute these intricate spatial configurations into a format that the LLM can interpret: by translating the map into a two-dimensional matrix structure. In this formulation, mineral fields are represented as elements within a matrix, interconnected to form an undirected network (see Fig. 4 and Fig. 5). This matrix contains the relational and locational data of the battlefield, which the LLM can process. Through this translation into a matrix, which effectively maintains the spatial relationships and proximities of in-game elements, we enable the LLM to analyze and comprehend the spatial dynamics of StarCraft II terrains.

\subsubsection{The SC2 Brain}

The function of the SC2 Brain is to translate natural language-based tactical concepts, as inferred by the Overmind Brain in response to environmental conditions, into executable commands within the StarCraft II environment. The advantage of this system lies in its capacity to prevent LLM from simultaneously generating intricate, high-dimensional tactical directives and detailed low-dimensional instructions in a single inferential step. By disentangling the complexity inherent to strategic forethought from the specificities required for in-game execution, the SC2 Brain ensures the operational clarity and efficiency required to execute Overmind Brain commands with accuracy. This approach to command translation facilitates a more nuanced and responsive adaptation to the rapidly evolving landscape of competitive gaming strategies.
The input information of the SC2 Brain consists of several parts:\\
(1) \textbf{Tasks with act as a Zerg player}, such as “\textit{You are a professional Zerg player in StarCraft II. You know all the dependencies between Zerg units, Zerg buildings, and Zerg technological research...}.”\\
(2) \textbf{The strategy that needs to be translated}, which comes from the output of the Overmind Brain.\\
(3) \textbf{Response format}, which aims to optimize the efficiency with which generated commands are executed within the StarCraft II environment.
For instance, with the command below:\\
“(Overlord, A1)->(Move)->(B1)”\\
In this command structure, the first set of parentheses identifies the Zerg unit requiring manipulation. The second set denotes the type of operation to be conducted, such as spawning or moving. The third set specifies the target location of the operation, which, in the case of a move command, corresponds to the destination point (e.g., B1, B2, etc.).\\
This syntax facilitates precise control and coordination of game units, enabling complex strategic maneuvers to be broken down into a sequence of simplified actionable steps. By standardizing the command framework in this manner, the execution becomes more accessible and allows for a higher degree of fidelity in translating strategic intents into in-game actions.\\
Given that our own units retrieved from self.unit are each assigned a unique tag identifier, we have experimented with directing the LLM to individually manipulate each distinct unit, as demonstrated below:\\
“*(Unit(name='Zergling', tag=4362338305))->(Attack)->(B1)”\\
“*(Unit(name='Zergling', tag=4361551873))->(Attack)->(B1)”\\
“*(Unit(name='Zergling', tag=4359454722))->(Attack)->(B1)”\\
…\\
However, we observed that the LLM's capability to adeptly control disparate units was suboptimal. Moreover, in situations necessitating the deployment of significant military forces, micromanagement of single units proved to be superfluous. As a result, our current methodology for our assault units, such as Zerglings, involves the issuance of collective commands—employing a notation like "(Zergling, A1)->(Attack)->(B1)"—to dispatch a group of Zerglings from location A1 to execute an attack on target B1, thus circumventing the need for individual assignment of combat tasks to each Zergling unit.\\

\subsubsection{The Command Center}

The Command Center serves a pivotal role within the StarCraft II gaming environment by translating commands interpreted by the SC2 Brain into actionable operations. This conversion process is executed through two integral components:

1) \textbf{Command Decoding}: This function is responsible for parsing the structured operational commands yielded by the SC2 Brain. Utilizing regular expressions, it identifies and extracts essential elements of each command, including the unit involved, the action to be performed, and the targeted location for the action to occur. It is noteworthy that when it comes to Zerg research commands, such as the "Metabolic Boost", there is no requirement for a target location, the specification of a target location is unnecessary, with the research name serving as the critical identifier.

2) \textbf{Condition Verification}: This component addresses occasional discrepancies in the understanding of build orders inherent to the LLMs for StarCraft II. On occasion, these LLMs may issue commands for morphing or construction that are not feasible under current gameplay conditions. The role of the Condition Verification module is to suspend such impractical commands until all prerequisites for morphing or construction are met, thereby enabling the execution of the commands.

Together, these two components of the Command Center ensure a efficient translation of strategic decisions from the SC2 Brain into the battleground, optimizing gameplay and strategy execution within StarCraft II.

\subsection{Swarm ReflexNet}
To address the infeasibility of LLMs for RTS environment, we introduce the innovative concept of Swarm ReflexNet, a framework designed to empower the basic Zerg units with the capability to execute simplistic, automatic reflex actions. This approach enhances the strategic efficacy of the Swarm by embedding conditional reflex behaviors into these units, thereby it is unnecessary for LLM to generate complex and detailed commands. 
Next, illustrative examples of state machine for the Swarm ReflexNet, featuring representative Zerg units such as the Drone, Overlord, and Zergling are discussed below.

\textbf{Drone's ReflexNet}. As depicted in Fig.6, the state transitions of the Drone are showcased when confronted with varying scenarios. The Drone is characterized by three distinct states: Gather (the default state) state, Attack state, and Flee state. These states are interchangeable under three specific conditions—\textit{Condition A ()}, \textit{Condition G ()}, and \textit{Condition F ()}—which are defined as follows:

\textit{Condition A ()}: In instances where the Drone is under assault from enemy units and the offensive power of the enemies within its visual range is less than that of the proximate Drones—for instance, an opposing SCV or a group of no more than three enemy Marines—the Drone shifts into the Attack state. It is important to note that this transition from Gather state to Attack state does not mobilize all nearby Drones to engage the intruder. This measured response is to prevent the adverse effects of enemy harassment tactics, ensuring that only the closest Drones with a sufficient combined attack power greater than the enemy’s are deployed against the intruder.

\textit{Condition G ()}: When adversaries within the Drone's visual range have been neutralized or have withdrawn from sight, the Drone reverts to its Gather state.

\textit{Condition F ()}: Upon the appearance of enemy forces within the Drone's visual range with an attack power exceeding that of nearby Drones—such as a squad of more than three enemy Marines or a group of Hellions—the Drone will transition into the Flee state.

\begin{figure}[htbp]
  \centering
  \label{fig:fig3}
  \includegraphics[width=4.5in]{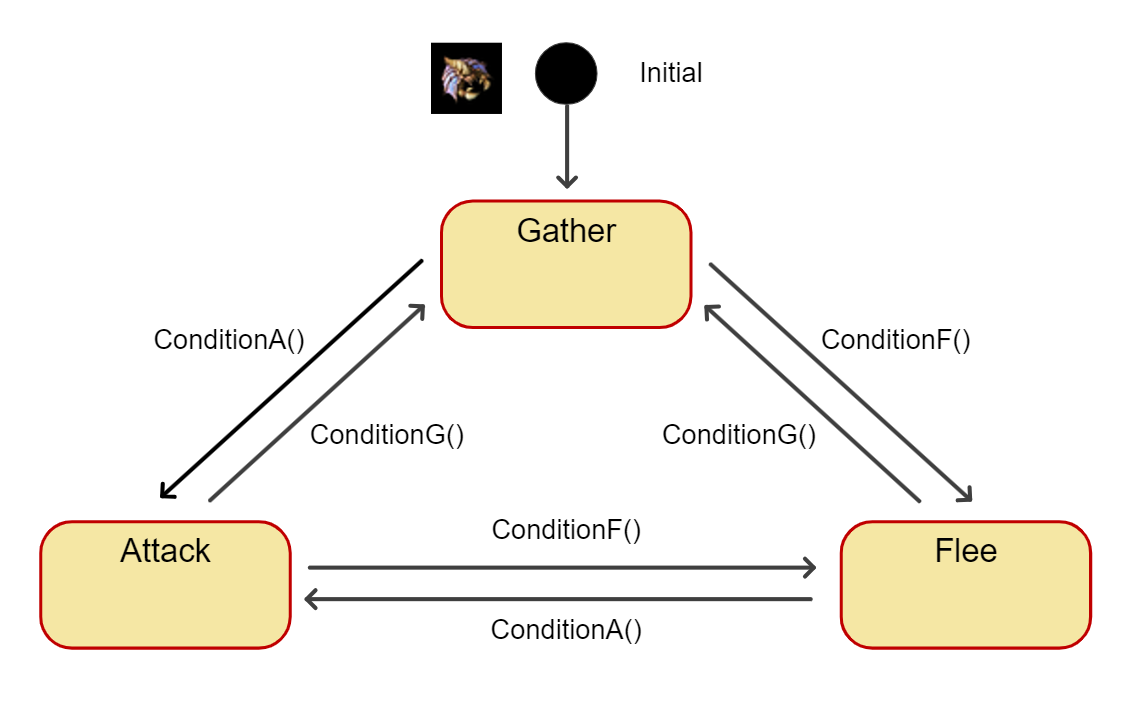}
  \caption{The state transitions of the Drone.}
\end{figure}

\textbf{Overlord's ReflexNet}. As shown in Fig. 7,  the state transitions of the Overlord is markedly simpler when compared to that of the Drone, with implementations intrinsic to the StarCraft II framework. The Overlord's state machine consists of two primary states: Idle state (the default state) and Flee state, which are governed by two specific conditions: \textit{Condition F ()} and \textit{Condition I ()}. These conditions are elucidated as follows:

\textit{Condition F ()}: When an Overlord is under attack, it will transition into the Flee state for self-preservation, initiating movement toward the nearest friendly unit or the edge of the map for safety purposes according to the situation.
While ordinary Overlords often eliminated by enemy forces before reaching the proximity of allies, this mechanism proves highly effective for its upgraded counterpart, the Overseer, which boasts superior mobility.

\textit{Condition I ()}: When hostilities cease or when hostile forces within the Overlord's visual range have been eliminated, the Overlord will transition back to the Idle state.

\begin{figure}[htbp]
  \centering
  \label{fig:fig3}
  \includegraphics[width=2.3in]{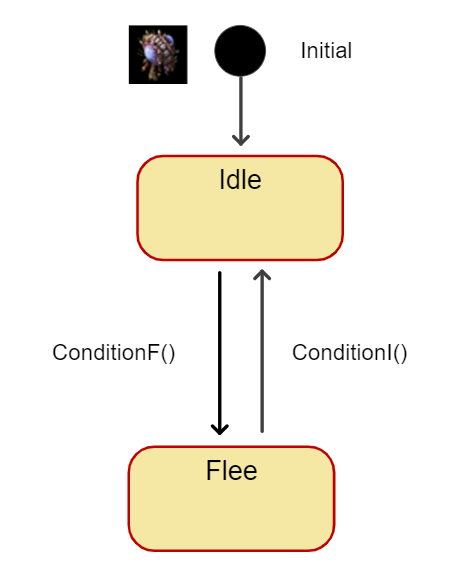}
  \caption{The state transitions of the Overlord.}
\end{figure}

\textbf{Combat unit's ReflexNet}. For Zerg's primary offensive units, such as Zergling, Roach, Hydralisk, etc., the state transition diagrams of Zerg combat units share similarities. Herein, the Zergling's state machine is shown in Fig. 8. The Zergling operates within three states: Idle state (the default state), Attack state, and Flee state, with transitions dictated by three distinct conditions: \textit{Condition A} (), \textit{Condition I} (), and \textit{Condition F} ().

\textit{Condition A} (): During the early stages of gameplay, when Zerglings are dispatched for aggressive maneuvers against an enemy base, they prioritize the assault on the opponent’s combat units, such as Marines, instead of attacking enemy buildings. After neutralizing these potential threats, Zerglings will proceed to reconnoiter the enemy's main mineral line, aiming to prioritize the elimination of enemy SCV and thereby disrupt the adversary's economic stability. During skirmishes, a Zergling group will enact tactical flanking maneuvers to initiate attacks from multiple angles, specifically targeting enemy units with high offensive capabilities but relatively low defensive fortitude, such as Siege Tanks. 
Similarly, in engagements where a mixed-army composition of Zerg units (comprising Zerglings, Roaches, and Ultralisks, which are limited to ground attacks, as well as Mutalisks and Hydralisks capable of targeting both ground and aerial units) confronts an Terran force that includes units such as Marines and Medivacs, the Mutalisks and Hydralisks are programmed with the tactical priority to first neutralize the Medivacs in order to disrupt their healing support for the Terran infantry.

\textit{Condition I} (): When Zerglings have finished their combat assignments or completed tasks generated by the LLM—i.e., all enemy units at the target location have been eliminated—the Zergling state transitions back to Idle. Similarly, the same strategy is applicable to other offensive units.

\textit{Condition F} (): When Zerglings operating outside the engagement zone or mineral field perimeter, particularly in smaller numbers (for example, around one to four Zerglings), when they’re under attack by enemy forces, they will prioritize retreat toward allied troop positions. This tactical principle can be analogously applied to other offensive units within the strategic framework.

This mechanism ensures that the Zerg units are capable of autonomously countering threats effectively, obviating the need for intervention by LLM, thereby bolstering their efficacy across diverse combat scenarios.

\begin{figure}[htbp]
  \centering
  \includegraphics[width=4.5in]{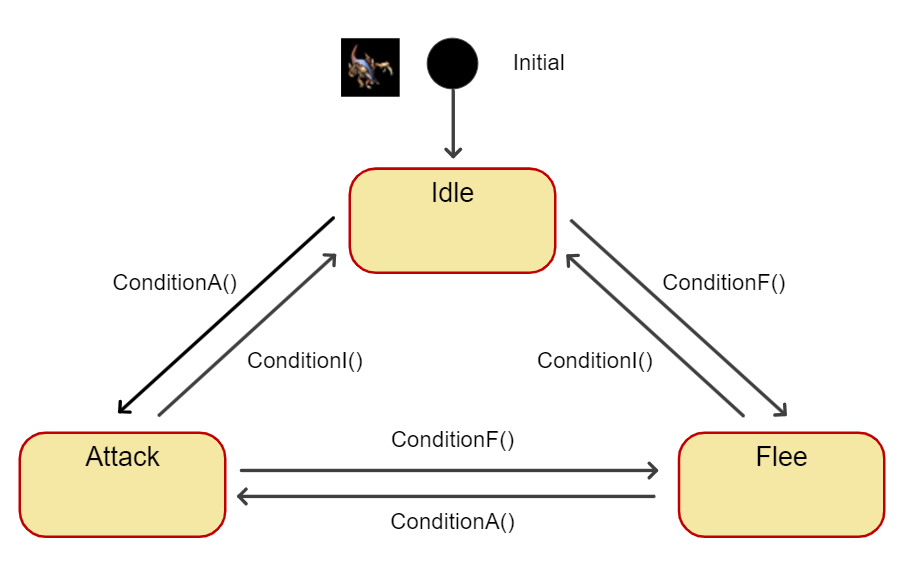}
  \caption{The state transitions of the Zergling.}
\end{figure}

\section{Experiment}
\subsection{Experimental Setup}
We utilize OpenAI’s gpt-3.5-turbo as the large language models of the Overmind Intelligence Matrix. Python-sc2 package is leveraged as the interaction of the SwarmBarin and StarCraft II environment. 

\subsection{Evaluation results}
To verify the effectiveness of the proposed method from diverse perspectives, we conducted a series of 30 experiments within a consistent experimental environment. The experiments were carried out on the professional competitive map "Automaton LE," where the SwarmBrain, played as the Zerg race, while the opponent, Computer, played as the Terran race. The Computer was set to compete at five distinct levels of difficulty, namely: Very Easy, Easy, Medium, Medium Hard, and Hard. Since the positions of both sides are random at the beginning of each match, we only start the game with the Zerg (SwarmBrain) in the lower left corner and the enemy Terran in the upper right corner.

\subsubsection{Win rate and average match time against Computers of varying difficulty levels}
Fig.9 (a) presents the number of victories secured by SwarmBrain in a series of 30 matches against Computers with five different difficulty settings. It can be observed that SwarmBrain consistently prevailed in all matches against Computers categorized as Very Easy, Easy, Medium, and Medium Hard. However, against the Hard difficulty Computer, SwarmBrain experienced a total of seven defeats. Analysis of these losses indicates that in three instances, commands issued by the core Overmind Intelligence Matrix powered by LLM were not correctly executed by the Python scripts. These scripts triage commands from the SC2 Brain using regular expressions and interface with the gaming environment through the python-sc2 library. An additional two defeats were the result of suboptimal strategic commands from the Overmind Intelligence Matrix, resulting in economic collapse. For example, the imprudent construction of an extra Hatchery at location B2 led to a sustained economic downturn for the Zerg, rendering them incapable of producing sufficient military units. This economic misstep contributed to the loss of nearly all defensive forces during the Computer's second wave of assault. In another two instances during the late game phase, the Overmind Intelligence Matrix showed a preference for producing a combined airforce composed of Mutalisks and a ground force consisting of Zerglings and Roaches for a coordinated strike at the enemy's B2 position. Unfortunately, these forces were inadequate against the opponent's composition of Siege Tanks, Marines, Medivacs, and Thors, which led to a significant reduction in population.

\begin{figure}[ht]
    \centering
    \begin{subfigure}[b]{0.45\textwidth}
        \centering
        \includegraphics[width=\textwidth]{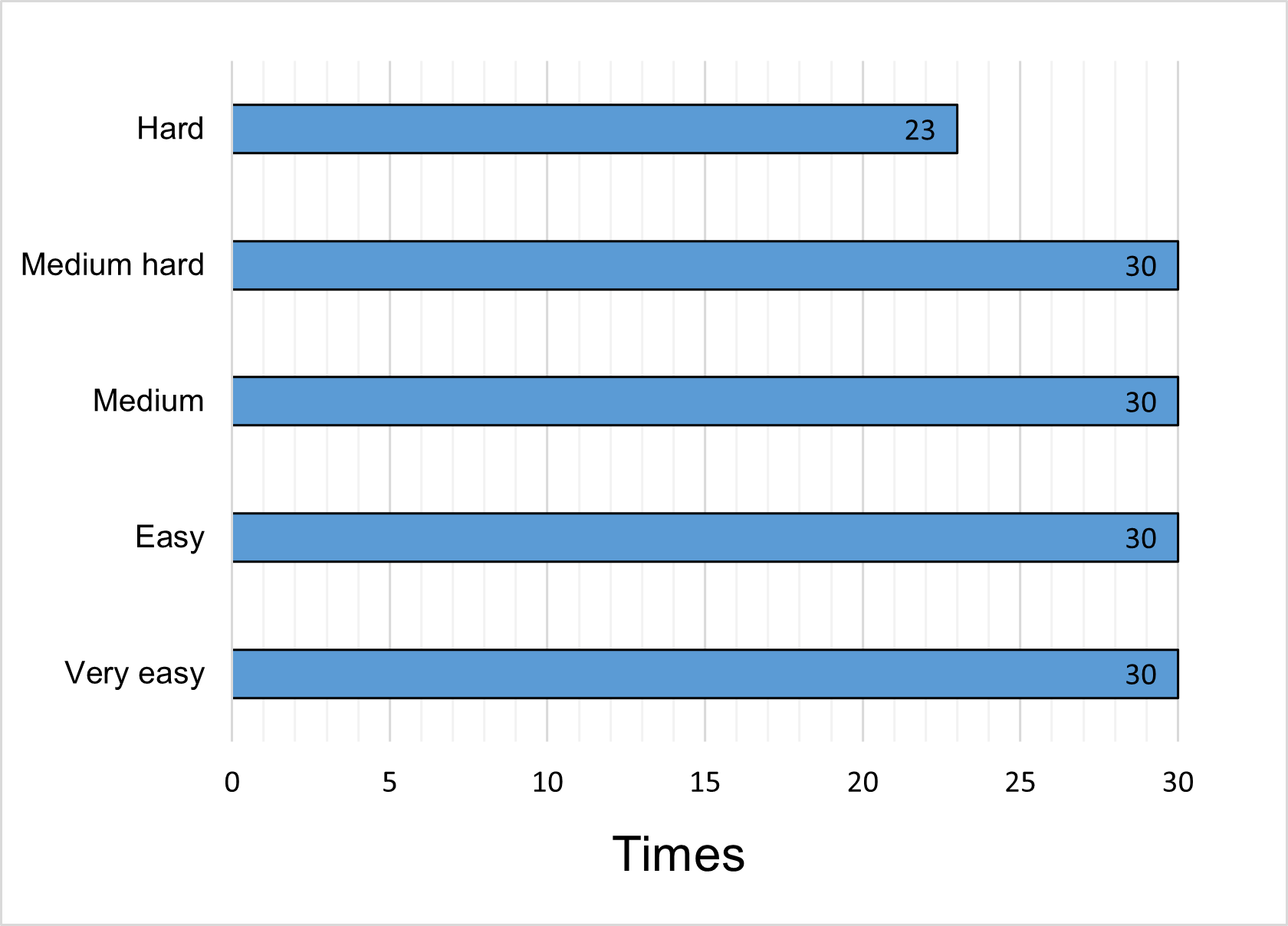}
        \caption{The number of victories secured by SwarmBrain against Computers with five different difficulty settings across 30 matches}
        \label{fig:pic1}
    \end{subfigure}
    \hfill 
    \begin{subfigure}[b]{0.45\textwidth}
        \centering
        \includegraphics[width=\textwidth]{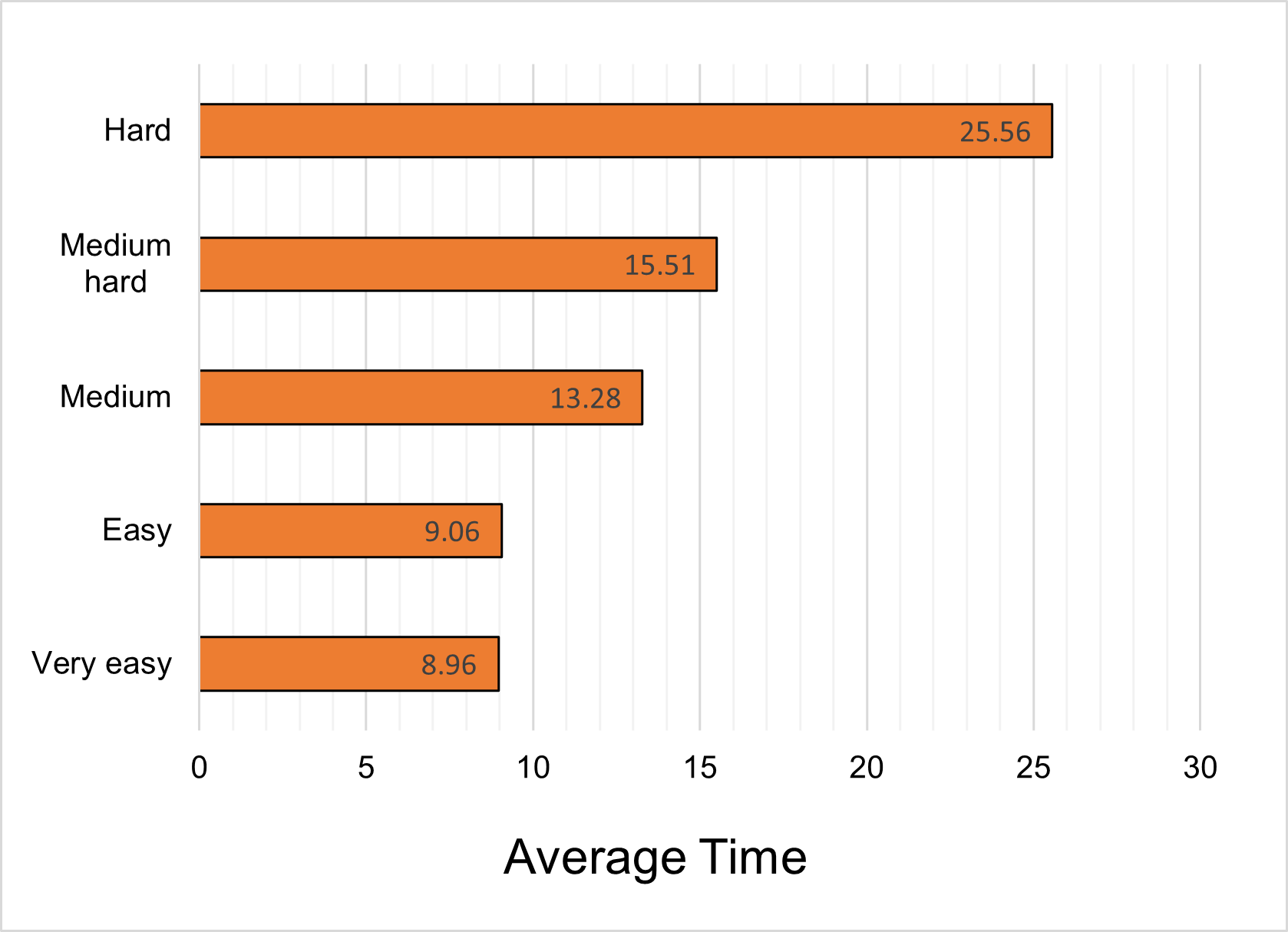}
        \caption{The average time to victory for SwarmBrain against Computers with five different difficulty settings across 30 matches}
        \label{fig:pic2}
    \end{subfigure}
    \caption{The win rate and average match time against Computers of varying difficulty levels}
    \label{fig:combined}
\end{figure}

\begin{figure}[htbp]
  \centering
  \begin{subfigure}[b]{\textwidth}
    \centering
    \includegraphics[width=4.5in]{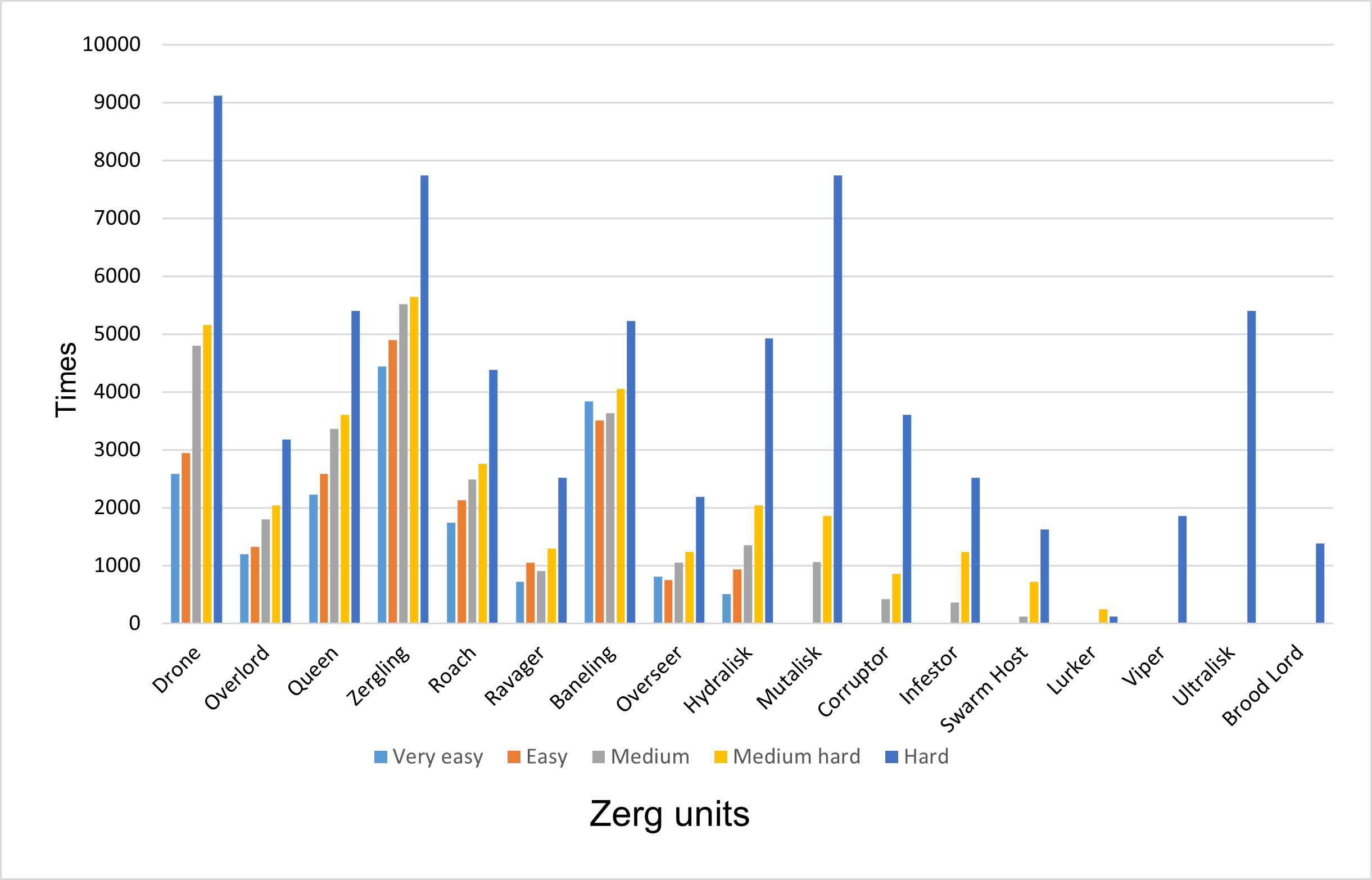}
    \caption{Frequency of command instructions for different Zerg units by SwarmBrain across 30 matches against Computers with five different difficulty settings.}
  \end{subfigure}
  
  \vspace{1cm} 
  
  \begin{subfigure}[b]{\textwidth}
    \centering
    \includegraphics[width=4.5in]{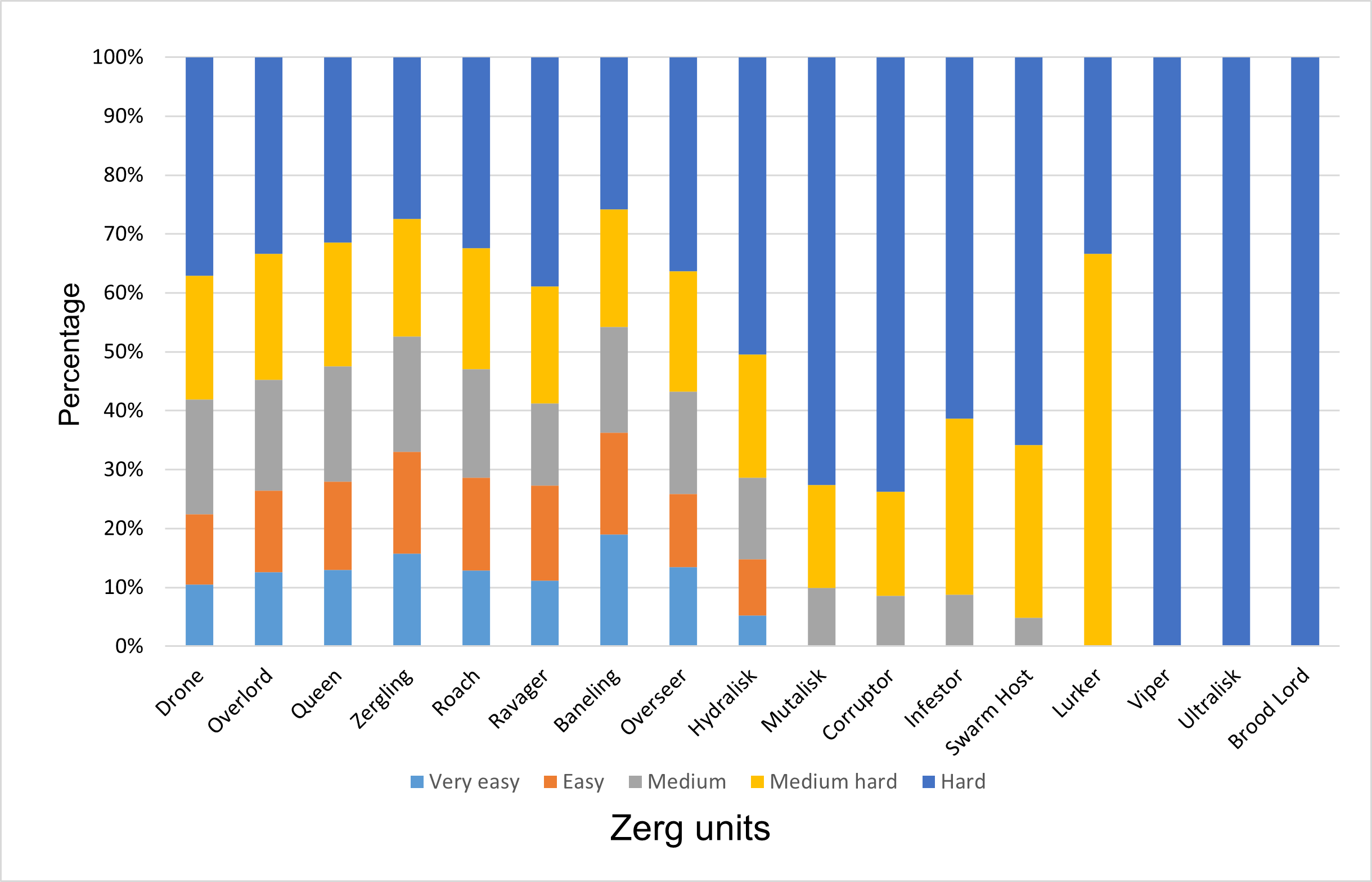}
    \caption{Percentage distribution of command instructions for different Zerg units by SwarmBrain across 30 matches against Computers with five different difficulty settings.}
  \end{subfigure}
  
  \caption{The ablation study on Swarm ReflexNet.}
  \label{fig:fig3}
\end{figure}

Fig.9 (b) represents the average time to victory for SwarmBrain competing against Computers with five different difficulty settings across 30 matches. It can be indicated that against Very Easy and Easy difficulty Computers, SwarmBrain achieved victory within an average duration of approximately 9 minutes. These swift conquests are typically attributed to the effective harass strategies executed by SwarmBrain's deployment of two waves of Zergling, Baneling, or Roach units, which essentially neutralized the Computer's offensive capabilities. It is noteworthy that in two matches, the army sent by SwarmBrain, while on the way to confront the Terran forces at target location B1 via waypoint B2, failed to detect and engage enemy units at B2 due to the 'Fog of War.' This oversight contributed to the prolongation of game duration. In matches against Medium and Medium Hard difficulty Computers, average victory times increased in comparison to those obtained against Very Easy and Easy computers. The reasons for the increase in match duration were multifaceted, involving complex factors related to the overall battlefield dynamics. Against the Hard difficulty computer, average victory duration peaked, exceeding 25 minutes. This significant increase is reflective of the increased strategic challenge and resilience posed by high-level Computer opponents, requiring more nuanced and adaptive gameplay to secure victory.

\subsubsection{Command analysis during SwarmBrain matches against Computers of varying difficulty levels}

Fig.10 (a) represents the frequency of command instructions (including both training and attack commands) for different Zerg units by SwarmBrain across 30 matches against Computers with five different difficulty settings. Furthermore, Fig. 10 (b) shows the percentage distribution of Zerg unit types employed against Computers with five different difficulty settings. Observational analysis reveals that when confronting Very Easy and Easy difficulty Computers, SwarmBrain exhibits a propensity to deploy fundamental Zerg attacking units such as Zerglings, Banelings, and Roaches. Higher-tier units designed for late-game engagement, such as Mutalisks, Infestors, and Ultralisks, are conspicuously absent from these match dynamics. This unit selection trend stems from the observation that SwarmBrain typically approaches the brink of victory following the first wave of attacks by the combination of Zerglings, Banelings, and Roaches, thereby there is no chance to train and produce more advanced Zerg units. A similar tactical approach applies to matches against Medium and Medium Hard difficulty Computers. However, as the match durations extend and the adversary's attacking unit composition evolves, SwarmBrain seizes the opportunity to issue commands for training more robust units such as Hydralisks, Mutalisks, Corruptors, Swarm Hosts, and Ultralisks. In scenarios involving the Hard difficulty Computer, the SwarmBrain's command instructions for Zerg units is notably more comprehensive. Despite the diverse unit options, SwarmBrain demonstrates a reluctance to utilize Lurkers, favoring instead a combined attack strategy incorporating both ground forces and aerial forces units. The chosen ground forces consist of Zerglings, Banelings, Roaches, Ravagers, and Ultralisks, while the aerial forces comprises Overseers, Mutalisks, Corruptors, and Brood Lords. This strategic predilection for integrative assaults reflects SwarmBrain’s commitment to undertake more diverse and all-encompassing offensives.

\subsubsection{Discussion on the SwarmBrain strategy}
\textbf{SwarmBrain's scout intelligence.} Within both contemporary and historical contexts of warfare, intelligence has invariably played a pivotal role in determining the outcome of conflicts. Recognizing the paramount significance of timely and accurate battlefield information, SwarmBrain diligently employs Overlords or Zerglings to surveil the opponent’s mineral fields frequently, furnishing itself with up-to-the-minute intelligence on enemy movements and strategies.
At the game’s begining, with the intent to garner expansive intelligence and gain insights into the adversary’s activities, SwarmBrain strategically dispatches Overlords to scrutinize the opposing player’s primary (B1) or secondary mineral fields (B2). Fig. 11 illustrates the Overlord sent by SwarmBrain embarking on a reconnaissance mission toward the enemy’s B2 mineral field, aptly showcasing SwarmBrain’s commitment to initial intelligence gathering.

\begin{figure}[htbp]
  \centering
  \includegraphics[width=5.0in]{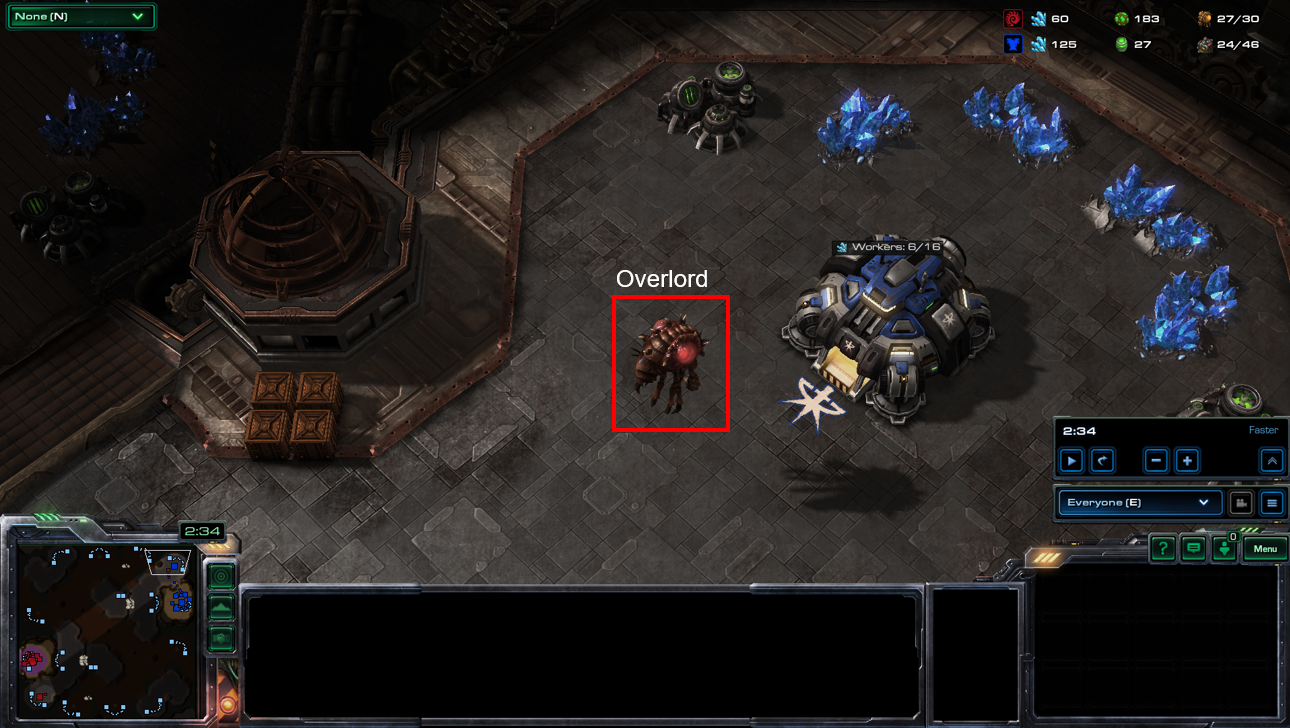}
  \caption{Overlord dispatched by SwarmBrain en route to conduct reconnaissance on the B2 mineral field.}
\end{figure}

\textbf{Early stage of the match.} Upon detecting a tangible threat to the A2 location from enemy forces—a situation classified as Critical Battlefield Information—SwarmBrain responds swiftly. 
Roaches in idle state alongside idle Zerglings from A2, are promptly mobilized under the guidance of Overmind Brain and Swarm ReflexNet to repel the encroaching threat. Fig. 12 vividly captures SwarmBrain’s strategic response and decision-making in the face of an assault.
Following an earlier offensive in which the dispatched Zerglings and Roaches suffered extensive losses, SwarmBrain promptly dispatches Hydralisks from A1 to reinforcement A2. Fig. 13 depicts the ensuing engagement at location A2, where a robust collective of Hydralisks confronts the adversarial forces.

\begin{figure}[htbp]
  \centering
  \includegraphics[width=5.0in]{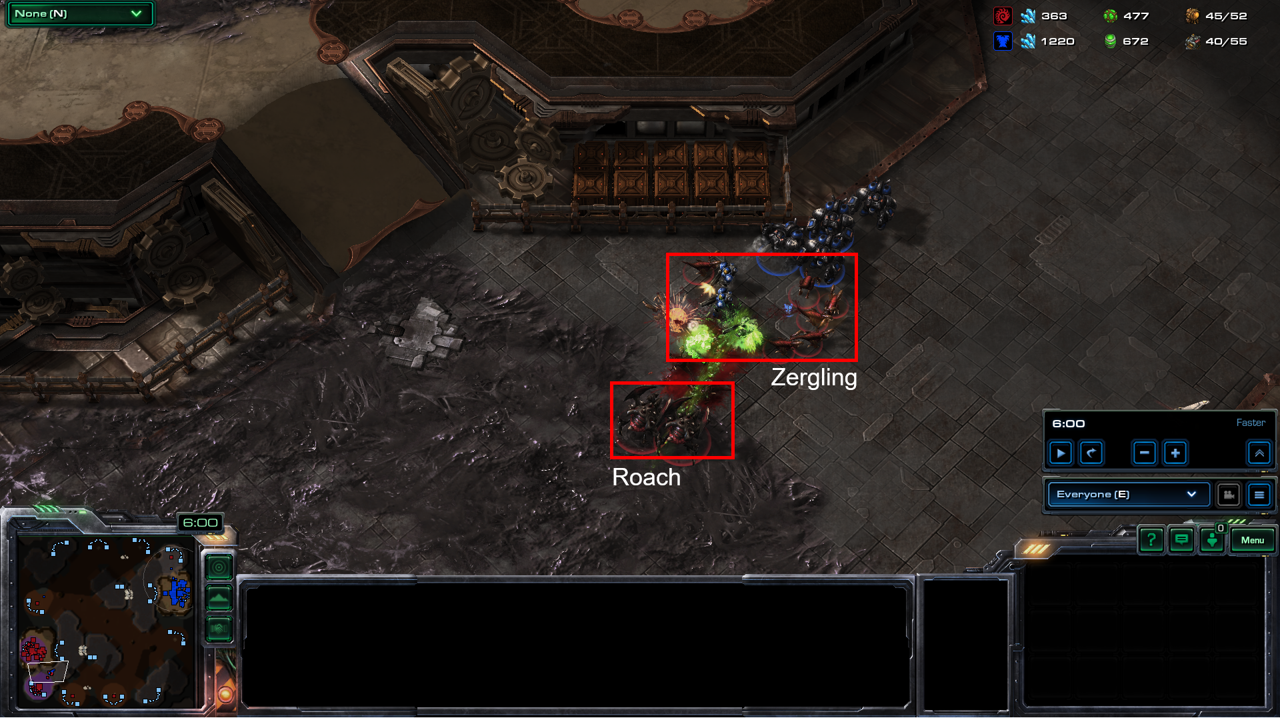}
  \caption{SwarmBrain's strategic response to an inbound attack.}
\end{figure}

\begin{figure}[htbp]
  \centering
  \includegraphics[width=5.0in]{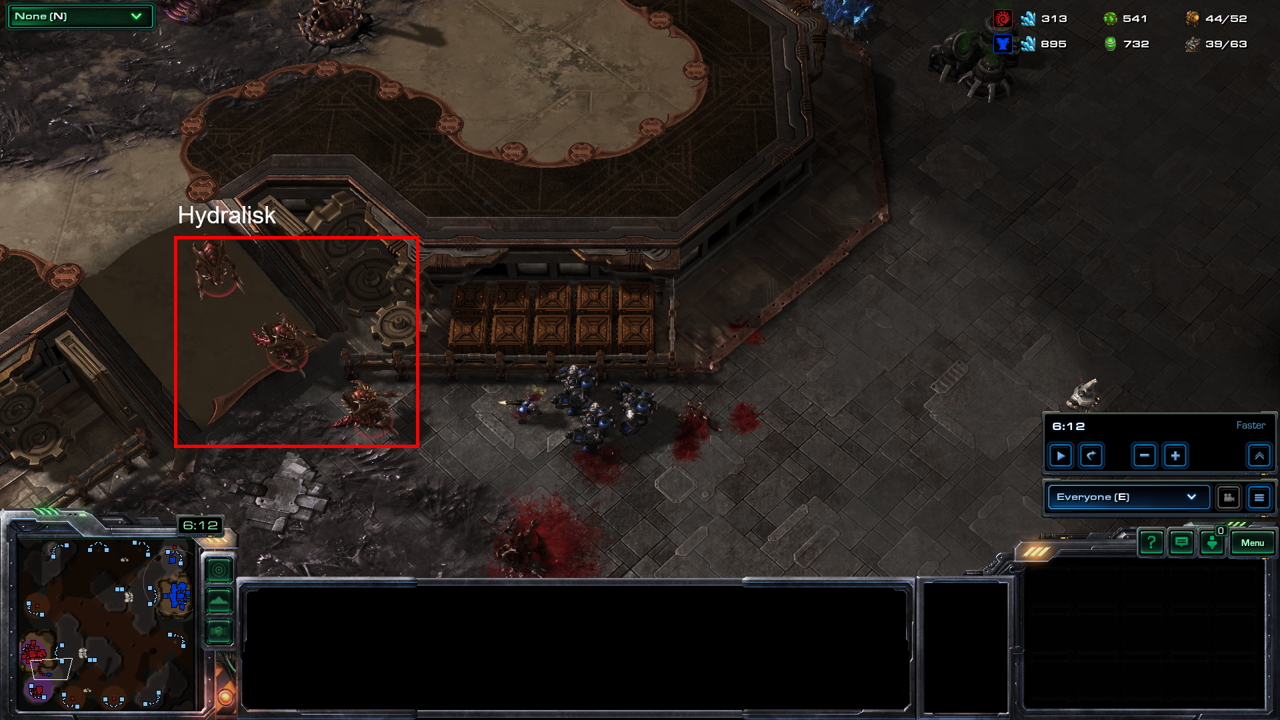}
  \caption{SwarmBrain opts for dispatching mightier Hydralisk units to reinforcement the battlefield.}
\end{figure}

\textbf{Later stage of the match.} As the competition advances into its later stages, SwarmBrain displays a marked preference for an air-ground troop amalgamation, complemented by Overseer reconnaissance. The ground forces are comprised of a few Zerglings, many Roaches, Hydralisks, and Ultralisks, while the aerial contingent consists of Mutalisks, Corruptors, Overseers, and Brood Lords. Fig. 14 showcases SwarmBrain’s preference for an air-ground troop amalgamation.

\begin{figure}[htbp]
  \centering
  \includegraphics[width=5.0in]{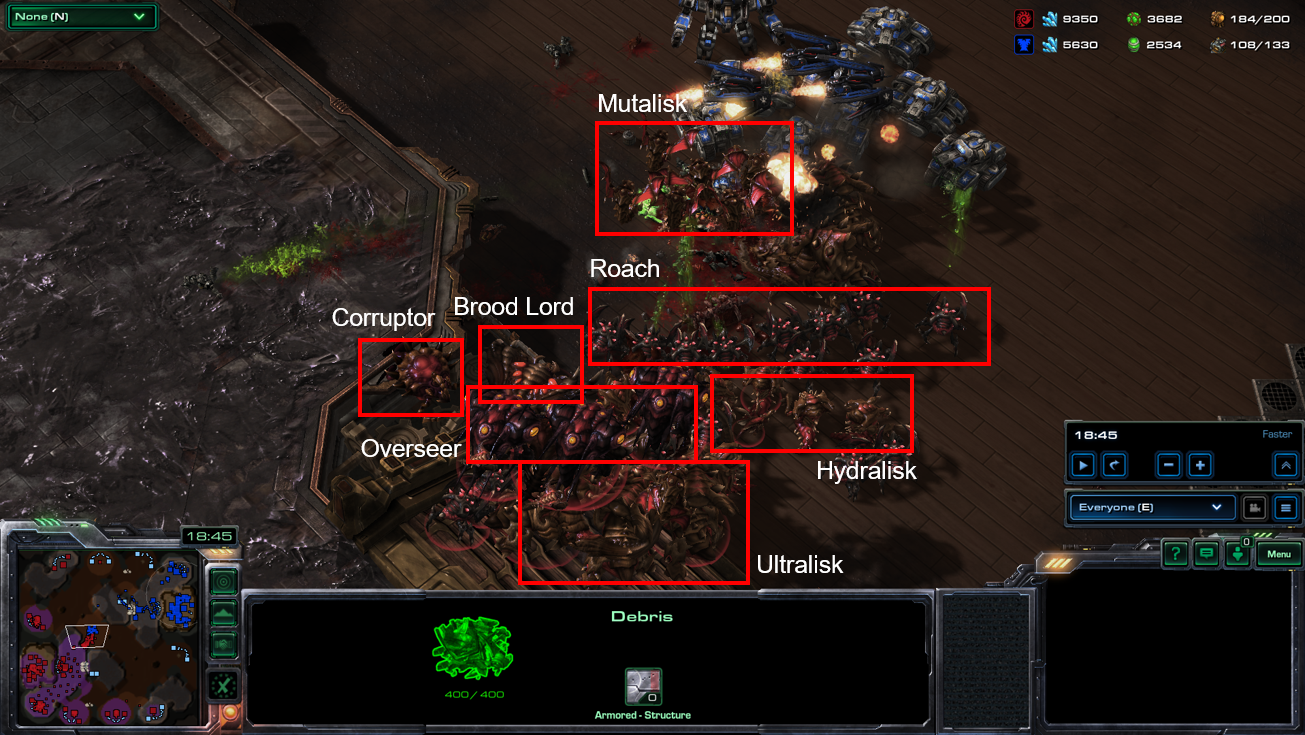}
  \caption{SwarmBrain's late-game military composition preferences.}
\end{figure}

After engaging in combat, SwarmBrain’s ground forces incurred a modest depletion of less than twenty supply. In response, SwarmBrain recalibrates its strategy and promptly directs the proximate A4 hatchery to commence extensive spawning of the powerful Ultralisk, aimed at assaulting the B1 location to replenish its forces. Fig. 15 illustrates SwarmBrain issuing commands for the prolific spawning of Ultralisks.

\begin{figure}[htbp]
  \centering
  \includegraphics[width=5.0in]{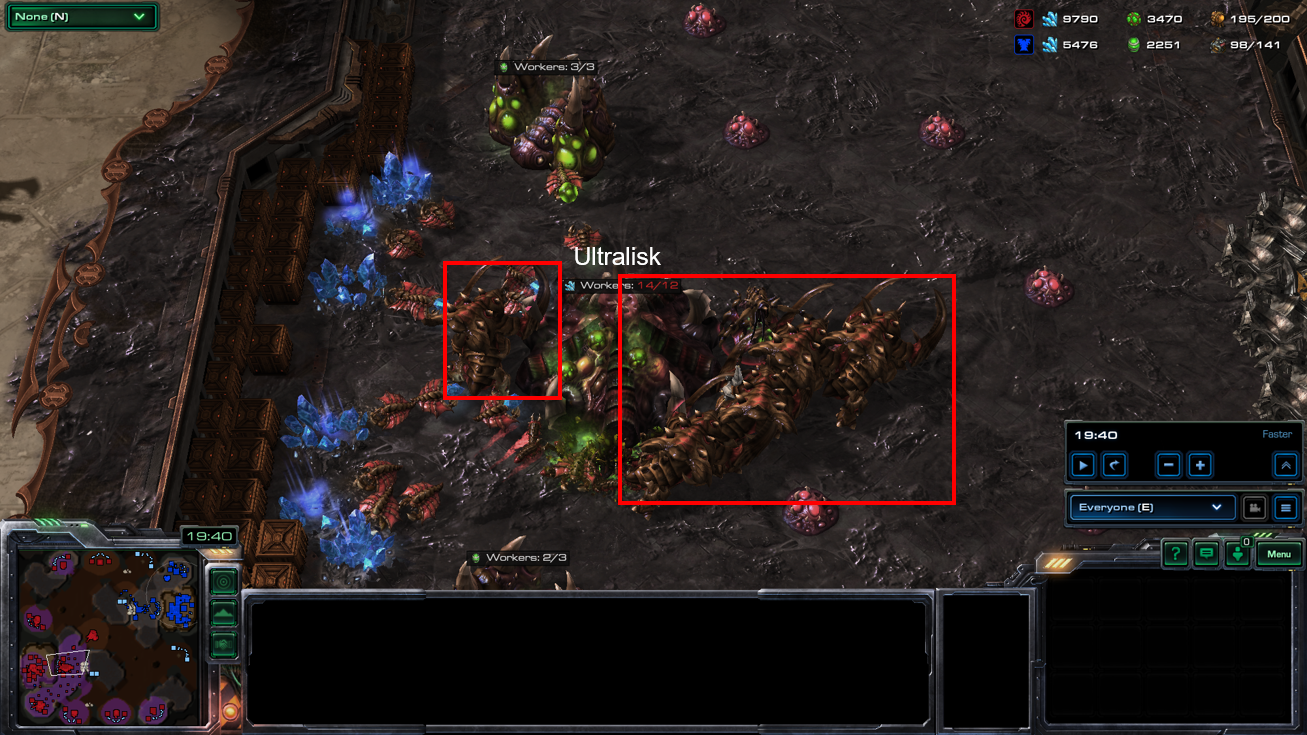}
  \caption{SwarmBrain spawns additional Ultralisks for replenishing its troop.}
\end{figure}

\subsection{Ablation study}

\textbf{Ablation study on gpt-3.5-turbo and gpt-4.5-turbo}.
Given that the Overmind Intelligence Matrix is LLM-based framework, we evaluated the performance differences when utilizing gpt-3.5-turbo versus gpt-4.0-turbo. Due to the significantly longer inference times of gpt-4.0-turbo compared to gpt-3.5-turbo, we devised two experimental conditions: the first scenario involved the Overmind Brain using the gpt-3.5-turbo, concurrent with SC2 Brain utilizing gpt-3.5-turbo. The second scenario combined the Overmind Brain using the gpt-4.0-turbo with SC2 Brain still employing gpt-3.5-turbo. Note that as the user of OpenAI continues to expand, there has been a corresponding increase in the inference time of GPT models. This development poses a significant challenge for leveraging LLM to secure victories in StarCraft II competitions, where the rapid processing capabilities of LLMs have played a crucial role in achieving success.

With gpt-4.0-turbo, we observed a deeper and more thorough understanding of game scenarios. However, the inference time of gpt-4.0-turbo was approximately two times longer than that of gpt-3.5-turbo. In the context of a per-minute gameplay environment, gpt-4.0-turbo was capable of conducting approximately three inferences, while gpt-3.5-turbo could perform six inferences, and with reduced Chain of Thought complexity, even more than twelve inferences. Although gpt-3.5-turbo presented a rapid inference capability, there were instances of positional misinterpretation of the game state, as depicted in Fig. 16. The figure indicates an erroneous inference instance wherein gpt-3.5-turbo issued a command for two Drones to construct two Extractors at point A8, where no Hatchery was present, instead of the correct point A4. 

Despite the more comprehensive and accurate output of gpt-4.0-turbo, the extended response times occasionally resulted in SwarmBrain’s failure to dispatch timely command orders to support Zerg units under attack, leading to defeat in some encounters. Therefore, considering the balance between inference speed and accuracy, we opted for gpt-3.5-turbo as the LLMs for both the Overmind Brain and SC2 Brain.

\begin{figure}[htbp]
  \centering
  \label{fig:fig3}
  \includegraphics[width=5.0in]{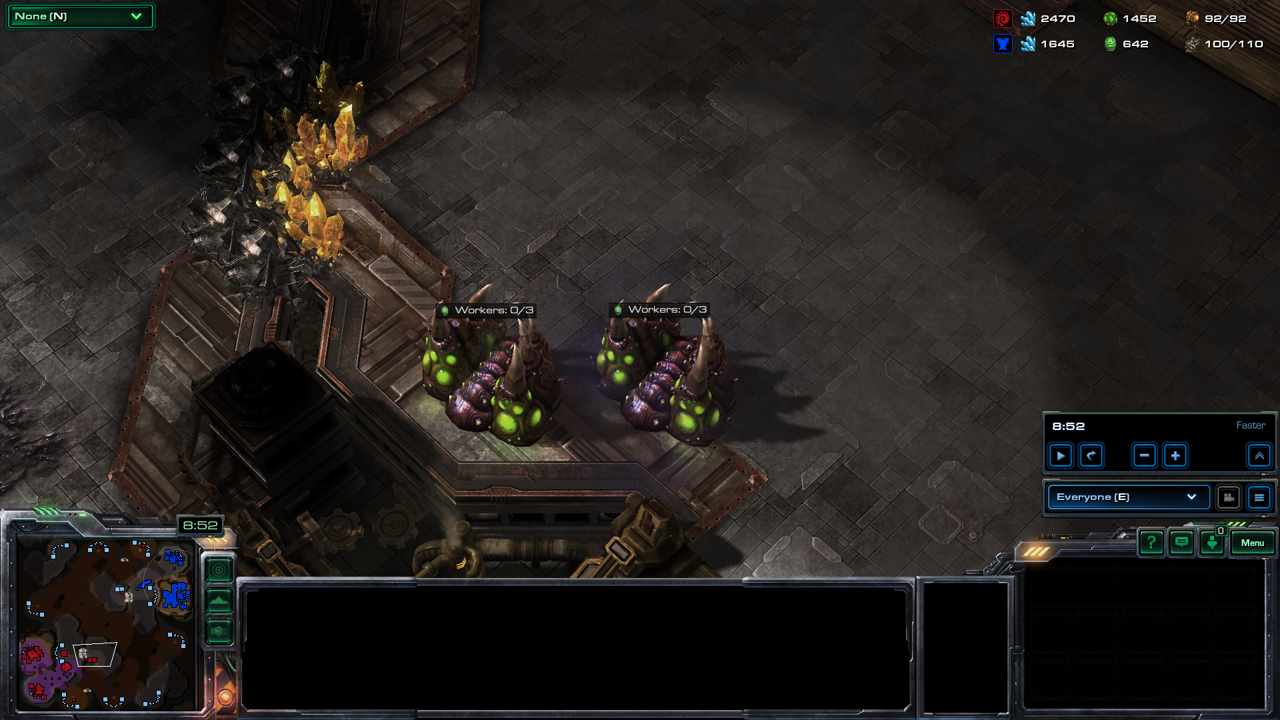}
  \caption{An erroneous inference example of gpt-3.5-turbo.}
\end{figure}

\textbf{Ablation study on Swarm ReflexNet}. 
Fig. 17 presents the experimental results of SwarmBrain with and without the Swarm ReflexNet. It can be observed from Fig. 17 (a) that in the absence of Swarm ReflexNet, when SwarmBrain dispatches Zerglings to attack the initial Terran base location, the Zerglings exhibit a propensity to target the Terran's supply depots and Barracks while neglecting the SCVs engaged in mining activities. Conversely, as depicted in Fig. 17 (b), the presence of Swarm ReflexNet redirects the Zergling's offensive prioritization towards the SCVs, which yields a substantial disruption to the opponent's economic.

\begin{figure}[htbp]
  \centering
  \begin{subfigure}[b]{\textwidth}
    \centering
    \includegraphics[width=5.0in]{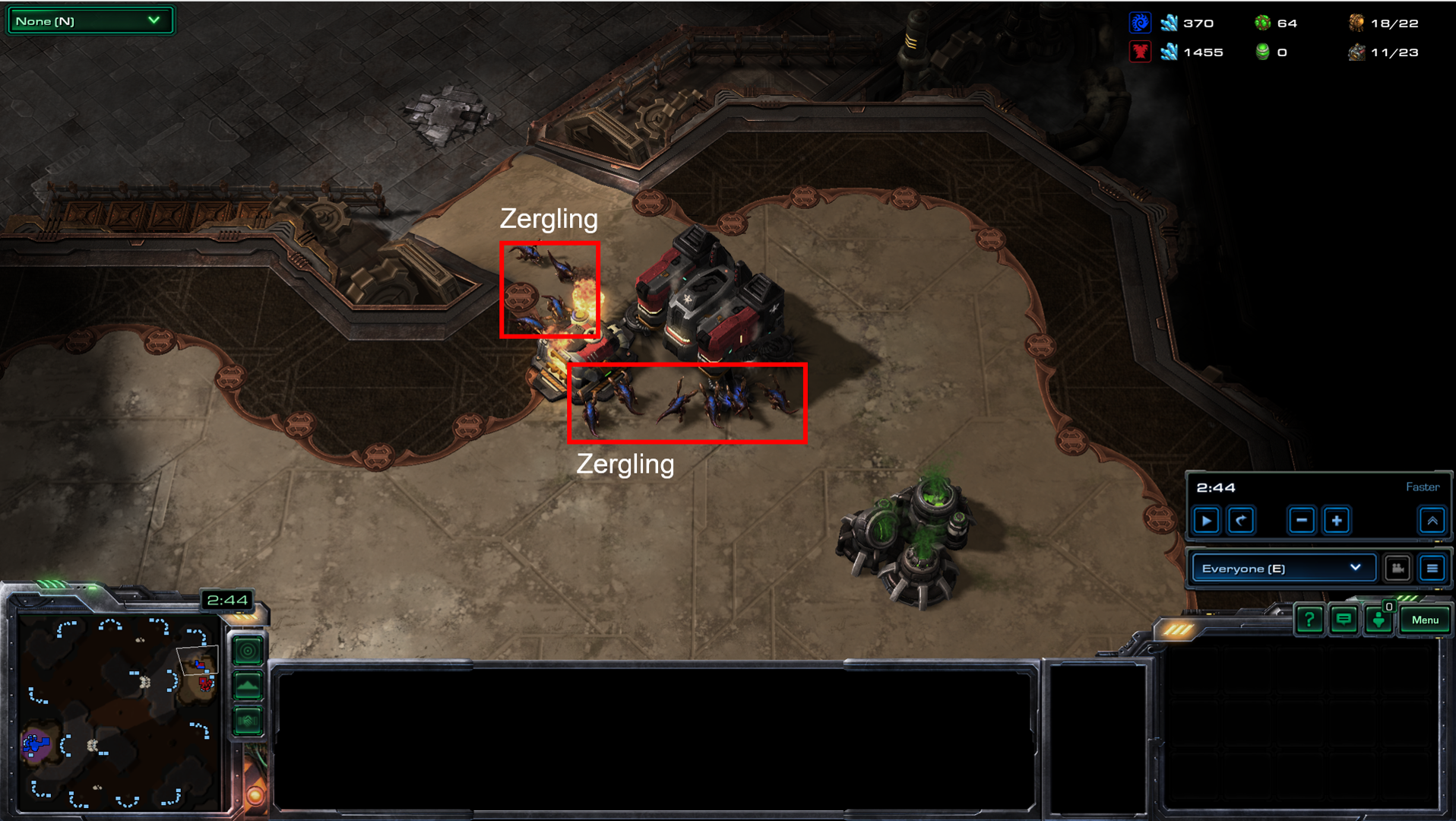}
    \caption{The Zergling's initial attack priority without Swarm ReflexNet.}
    \label{fig:fig3a}
  \end{subfigure}
  
  \vspace{1cm} 
  
  \begin{subfigure}[b]{\textwidth}
    \centering
    \includegraphics[width=5.0in]{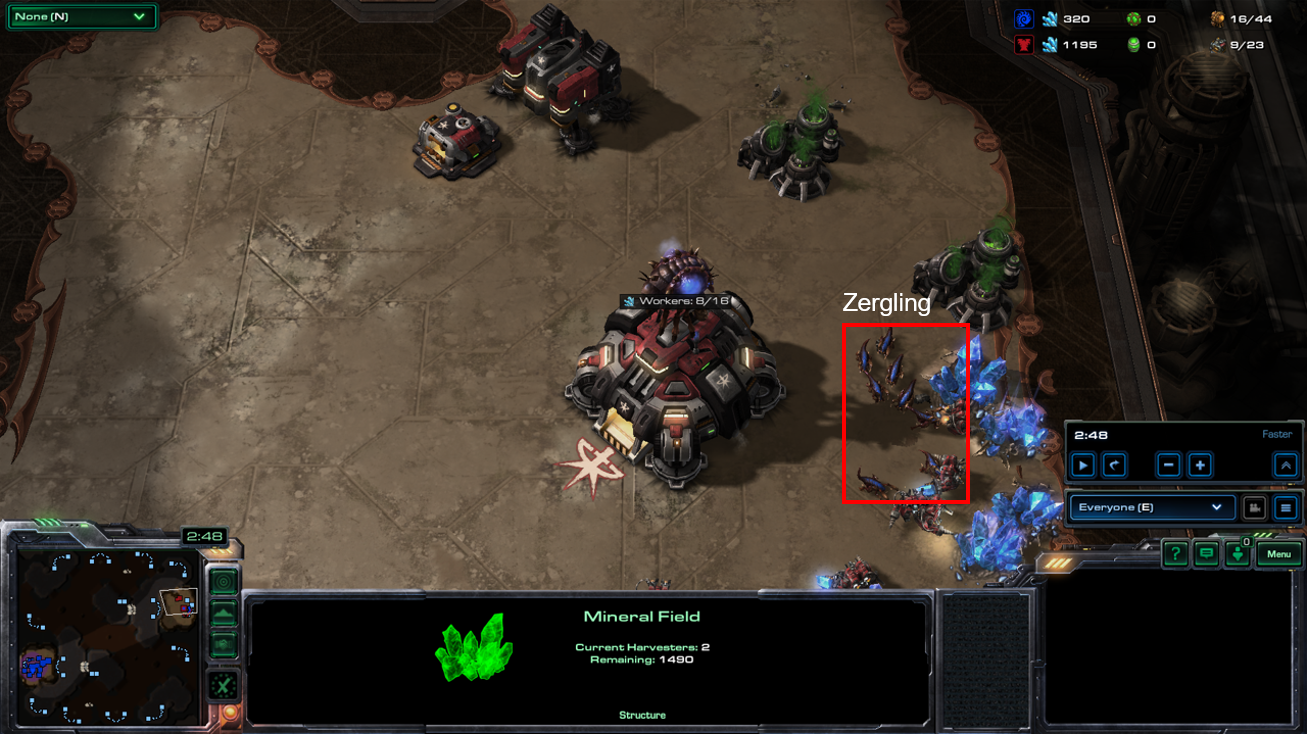}
    \caption{The Zergling's adjusted attack priority with Swarm ReflexNet.}
    \label{fig:fig3b}
  \end{subfigure}
  
  \caption{The ablation study on Swarm ReflexNet.}
  \label{fig:fig3}
\end{figure}

Fig. 18 illustrates the tactical engagements of the SwarmBrain’s Zerg forces during armed conflict. The Zerg army consists mainly of Zerglings and Roaches. In the depicted scenario, the Zerglings are programmed to prioritize flank maneuvers to assault the Terran Siege Tanks, leveraging their agility and close-combat prowess. Concurrently, the Roaches engage from a longer range, capitalizing on their robust ranged offensive capabilities to exert suppressive fire on the Terran army. This showcases the effectiveness of the Swarm ReflexNet.

\begin{figure}[htbp]
  \centering
  \label{fig:fig3}
  \includegraphics[width=5.0in]{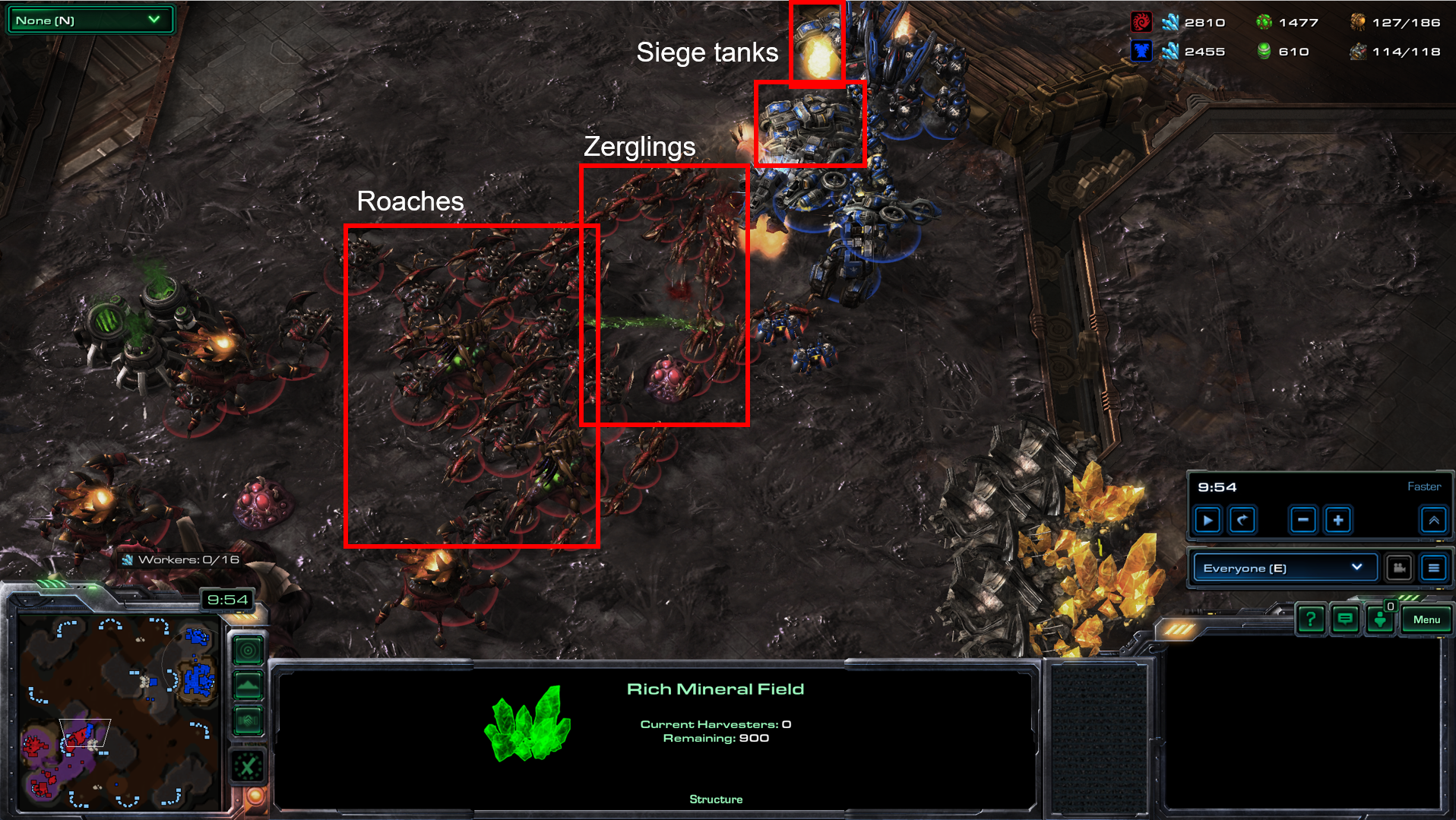}
  \caption{Prioritization of SwarmBrain's Zerg units during armed conflict.}
\end{figure}

\section{Conclusion}
In conclusion, our study presents SwarmBrain, an agent that uses large language models to efficiently perform real-time strategy tasks in StarCraft II. The SwarmBrain's design combines an Overmind Intelligence Matrix for high-level strategic planning and a Swarm ReflexNet for quick tactical reactions. The experiments demonstrate SwarmBrain's capabilities in resource management, territorial control, and tactical skirmishes, effectively defeating Computer-based opponents in most tested scenarios.

\section{Discussion}
In this paper, we introduce the SwarmBrain, an innovative approach designed for interaction with the StarCraft II gaming environment. While the SwarmBrain has achieved promising results, it still exists challenges that are not easily surmountable at the current stage. These challenging aspects present numerous avenues for future research and thoughtful consideration for advancement. These deficiencies are outlined as follows:

\textbf{Deficiency of Visual Information}: Predominantly relying on large language models for tactical reasoning through textual content, the SwarmBrain's comprehension of the battlefield is solely derived from text-transformed scenarios. This results in a lack of nuanced, two-dimensional spatial understanding. Although efforts have been made to provide a rudimentary grasp of environmental context through simple game map representation with map locational information, this process is still far from emulating human cognizance of spatial arrangement.

For instance, in StarCraft II, a common defensive tactic employed by human players, also referred to as a "wall-off" strategy, involves the strategic placement of Supply Depots and Barracks to blockade the entrance of a base, thereby preventing incursions from ground units. 
As illustrated in Fig. 19, this maneuver exemplifies a typical limitation encountered by the SwarmBrain. The figure clearly demonstrates the Terran players can halt a Zerg rush simply by positioning two Supply Depots and a Barrack at the entrance of the initial base, a scenario wherein SwarmBrain does not comprehend the full eradication of its Zergling assault due to its absence of visual comprehension of the placement of buildings.

\begin{figure}[htbp]
  \centering
  \label{fig:fig3}
  \includegraphics[width=5.0in]{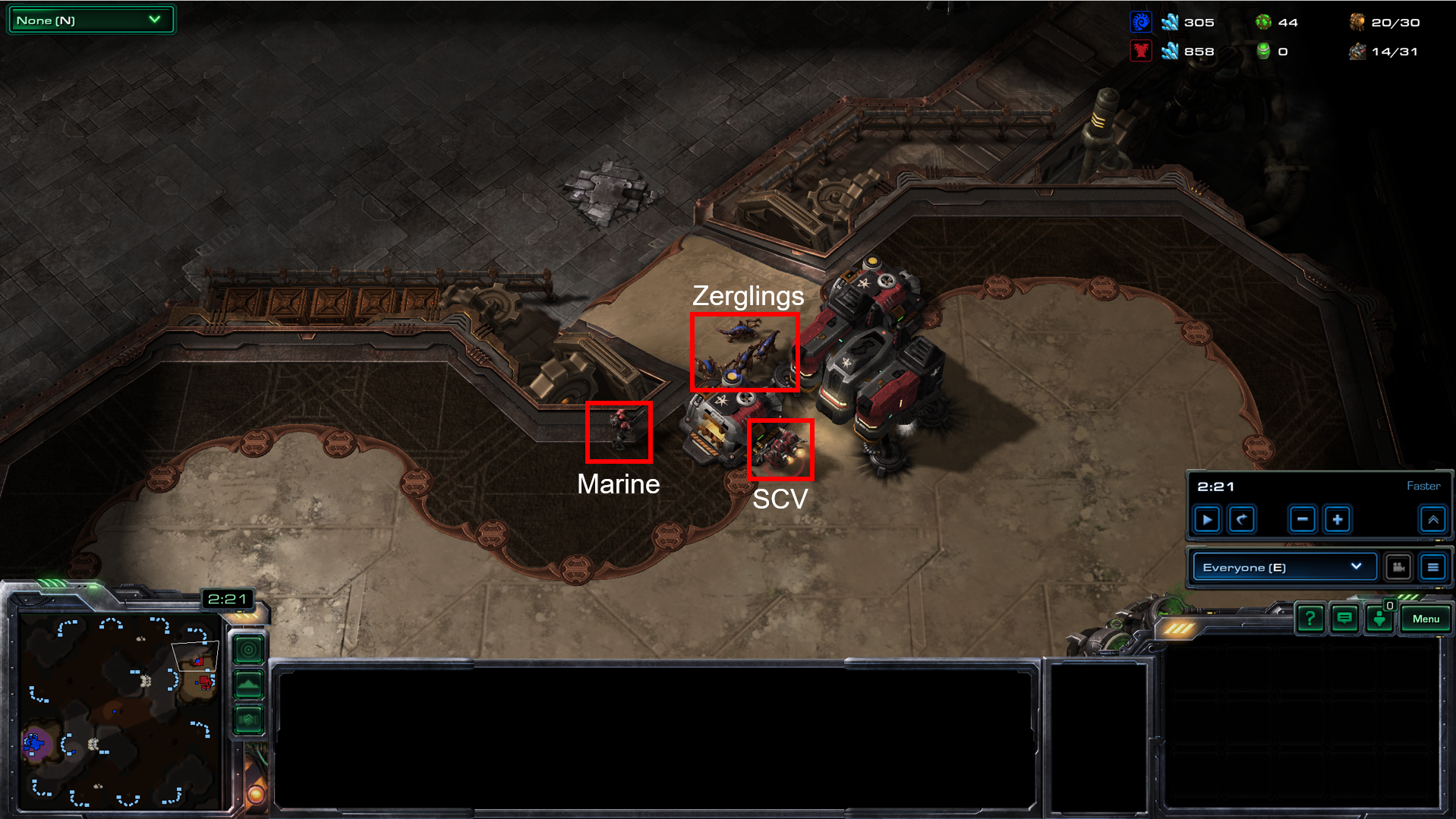}
  \caption{The human player's "wall-off" strategy defeat the SwarmBrain.}
\end{figure}

Furthermore, the precision in deploying the Queen's creep spread is hampered by the lack of visual feedback, often leading to suboptimal coverage that impedes the Zerg's offensive capabilities. In StarCraft II, professional Zerg players strive to maximize their Creep spread using as few Creep Tumors as possible, thus effectively extending their terran across the battlefield with minimal resource investment. 
Such a strategic positioning is crucial as it would expand the Zerg’s vision and mobility range, which are vital for territorial control and tactical advantages.
However, due to a lack of effective visual information processing, the SwarmBrain falls short in accurately and efficiently placing Creep Tumors. 
Currently, we employ a strategy that involves randomly spreading Creep within a circular area surrounding Zerg Hatcheries, or within the periphery of the existing Creep, for the deployment of subsequent Creep Tumors.
Instead, the Creep spread would erroneously progresses towards a location closer to the Hatchery sometimes. This has also resulted in a significant reduction in troop mobility, and ultimately cause in the defeat of the SwarmBrain contingent.

While multimodal models \cite{[gpt4v]} demonstrate exceptional ability to process both image and text data, integrating these with SwarmBrain within the StarCraft II environment proves inefficient. Given the inherent RTS nature of the game, the requisite reasoning time of the LLM negatively impacts the dynamic flow of gameplay. The added processing demands of multimodal models exacerbate this effect further, leading to elongated reasoning durations. Subsequent improvements contemplate mapping unit positional data onto a two-dimensional map, allowing the LLM to infer unit statuses based on their interrelations without significant latency.

\textbf{The Inference Velocity of LLMs}: The speed of reasoning exhibited by LLMs significantly affects gameplay outcomes, an issue previously deliberated. Despite the implementation of Swarm ReflexNet to accelerate proactive attack maneuvers, the latency stemming from awaiting LLM-deduced strategies is prominent. 
For example, when a match is launched, since the LLM needs the time to interface, SwarmBrain would needs 10 seconds pending LLM strategy formulation, resulting in substantial economic setbacks as only 12 Drones adjacent to the Hatchery continue resource collection. To compensate for the initial lag of approximately 10 seconds at the onset of gameplay, we currently initiate matches subsequent to the completion of inference by the LLM. This latency contributes to the current inability of SwarmBrain to rival advanced human players and more difficult computers. Given that the inference speed of LLMs is a constraint beyond our immediate capability to rectify, it remains one of the most pressing challenges to address.

\section*{Acknowledgments}
We would like to express our profound appreciation to the content creators on the YouTube channel specializing in StarCraft II AI videos. Their insightful tutorials and generous assistance played a crucial role in the advancement of our research. We are also immensely grateful to Zihan Li and Bin Dong, esteemed StarCraft II players, for their technical support regarding the methodologies presented in this paper, as well as for their invaluable feedback derived from competition with AI opponents. Their expertise contributed significantly to refining our strategies and enhancing the overall quality of our work.


\newpage
\section*{Author Profiles}

\begin{multicols}{2}
    
    \begin{wrapfigure}{l}{0.15\textwidth} 
    \includegraphics[width=0.15\textwidth]{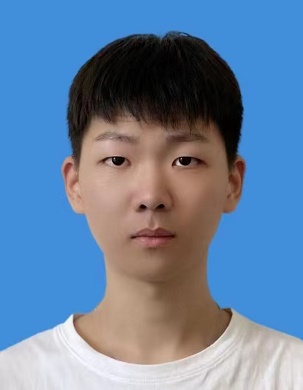} 
    \end{wrapfigure}
    
    \textbf{Xiao Shao} received the B.S. degree in computer science and technology from Nanjing University of Information Science and Technology, Nanjing, China, and the M.S. degree in software engineering from the Nanjing University of Information Science and Technology, Nanjing, China.
Alongside his academic pursuits, Xiao Shao is active in the field as an AI Researcher with BMW Archermind technology company. His professional role encompasses working on novel advancements in Natural Language Processing, Computer Vision, Deep Learning, Image Quality Assessment, and Large Language Model. Xiao Shao's academic and professional pathways are both geared towards leveraging artificial intelligence to solve real-world problems, blending theoretical knowledge with practical implementations. His work aims to bridge the gap between conceptual research and applied science in the fast-paced environment of AI technology development. His research endeavors are dedicated to the pursuit of truly autonomous Artificial General Intelligence.
Currently, he is the BMW Brilliance China AI Expert member.

    \begin{wrapfigure}{l}{0.15\textwidth} 
    \includegraphics[width=0.15\textwidth]{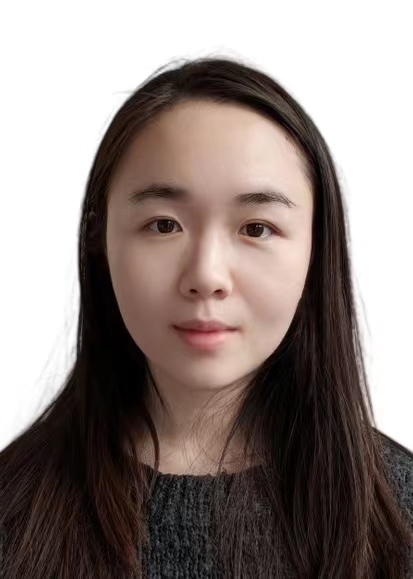} 
    \end{wrapfigure}
    
    \textbf{Weifu Jiang} received her B.Sc. in Financial Mathematics from the University of Sheffield, UK, and followed this by completing her M.Sc. in Computer Science at Cardiff University, UK. She currently holds a position as an AI Researcher at BMW Archermind Technology Company, where she specializes in the fields of Natural Language Processing and Reinforcement Learning. Weifu Jiang particularly focuses on developing advanced NLP and RL strategies to enhance the artificial intelligence capabilities in automotive technologies. Her work is pivotal in shaping the future of smart, intuitive interfaces and decision-making systems in vehicles. Her experience bridges the complex domains of finance, mathematics, and computer science, enabling a multifaceted approach to AI research and development. Currently, she is the BMW Brilliance China AI Expert member.

    \columnbreak 

    \begin{wrapfigure}{l}{0.15\textwidth} 
    \includegraphics[width=0.15\textwidth]{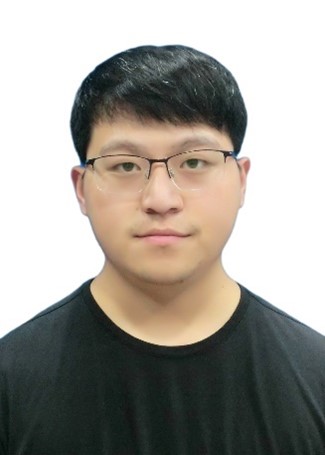} 
    \end{wrapfigure}
    
    \textbf{Fei Zuo} is currently undertaking a part-time Master's program in machine learning with a specialization in computer technology at East China Normal University, set to conclude in 2025. Starting in January 2023, he has been actively engaged in the development of large language model applications and the exploration of LLM-agent applications at BMW Archermind Technology Company.
    Fei Zuo's research pursuits encompass large language models, multi-modal large models, and the autonomous decision-making capabilities of LLM-agents. Currently, she is the BMW Brilliance China AI Expert member.

    \begin{wrapfigure}{l}{0.15\textwidth} 
    \includegraphics[width=0.15\textwidth]{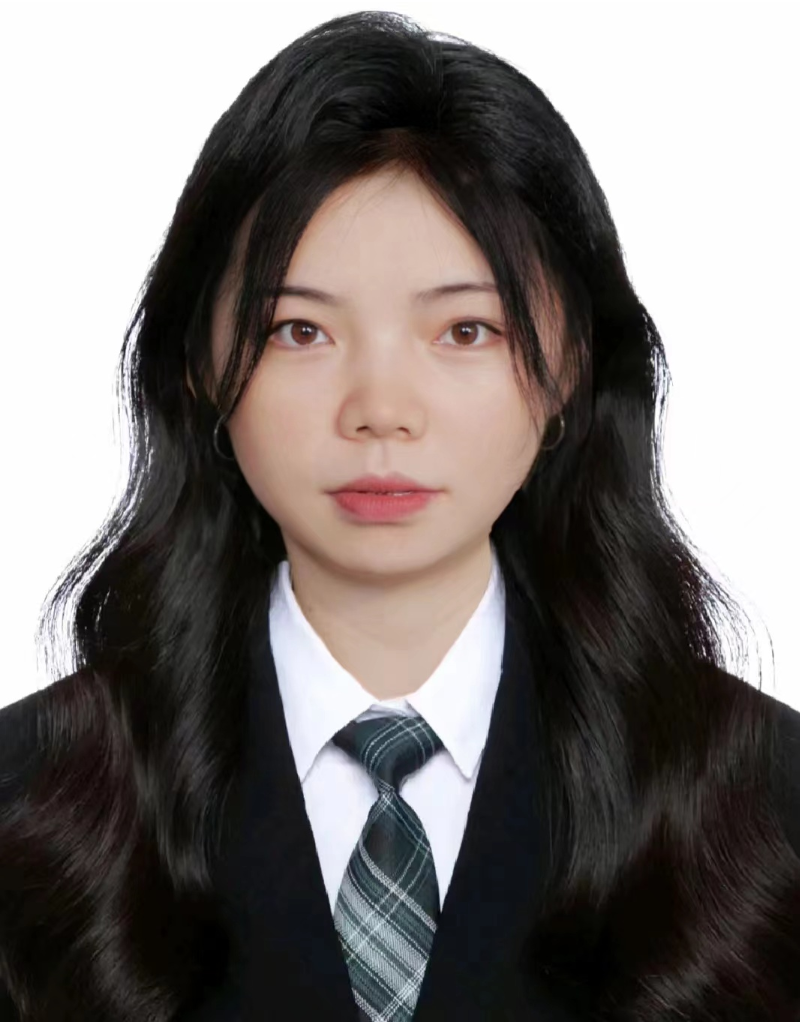} 
    \end{wrapfigure}

    \textbf{Mengqing Liu} received her B.Sc. in Internet of Things from Nanjing University of Information Science and Technology, Nanjing, China, and the M.S. degree in Computer Technology from Nanchang Hangkong University, Nanchang, China.
    After completing her graduate studies, Mengqing Liu embarked on her academic career in 2023, joining the School of Computer and Information Engineering at Nantong Institute of Technology, where she serves as an instructor specializing in Artificial Intelligence. Her research interests lie primarily in the fields of Computer Vision, Image Segmentation, Medical Image Processing, Natural Language Processing, and Large Language Models. She has been actively involved in advancing the state of research in these areas, aiming to contribute to both academic advancements and real-world applications.

\end{multicols}

\newpage

\section*{Appendix}
\subsection*{A. Prompt}

\subsection*{A.1 The Overmind Brain}
\subsection*{A.1.1 The components of the prompt for Overmind Brain}

The input prompt to gpt-3.5-turbo consists of several components:\\
(1) The task of the LLM.\\
(2) The map locational information.\\
(3) The repository of the strategies.\\
(4) Comprehensive battle assessment protocols.\\
(5) The current battle situation.\\
$\bullet$ \textbf{Critical battlefield information}:\\
For example:\\
Important!!!\\
Overlord have detected a group of Terran army is ready to attack the A4 and destory our army.\\
Zergling have detected a group of Terran army is assembling at B4.\\
...\\
...\\

$\bullet$ \textbf{Current Units}:\\
For example:\\
\{ \\
At point A1, there are: 12 Drone are gathering minerals in Hatchery, 3 Drone are gathering gas in Extractor1, 3 Drone are gathering gas in Extractor2, 11 Overlord are idling, 1 Queen constantly injecting eggs into Hatchery;

At point A2, there are: 15 Drone are gathering minerals in Hatchery, 3 Drone are gathering gas in Extractor1, 3 Drone are gathering gas in Extractor2, 2 Overlord are idling, 1 Queen constantly injecting eggs into Hatchery;

At point A3, there are: 10 Drone are gathering minerals in Hatchery, 3 Drone are gathering gas in Extractor1, 3 Drone are gathering gas in Extractor2;

At point A4, there are: 8 Drone are gathering minerals in Hatchery, 1 Queen constantly injecting eggs into Hatchery, 1 Queen are idling, 15 Roach are idling, 24 Zergling are idling, 2 Ravager are idling.

…\\
…\\
\} \\
$\bullet$ \textbf{Current Buildings}:\\
For example:\\
\{ \\
At point A1, there are: 1 Hatchery, 2 Extractor, 1 Spawning Pool, 1 Roach Warren, 1 Baneling Nest, 1 Evolution Chamber; 

At point A2, there are: 1 Hatchery, 2 Extractor, 1 Spawning Pool; 

At point A3, there are: 1 Hatchery, 2 Extractor;

At point A4, there are: 1 Hatchery, 2 Extractor;

…\\
…\\
\} \\
$\bullet$ \textbf{Current technological research}:\\
For example:\\
\{ \\
Metabolic Boost,\\
Melee Attacks Level 1,\\
Missile Attacks Level 1,\\
Ground Carapace Level 1,\\ 
…\\
…\\
\} \\
(5) Detected enemy Units.\\
$\bullet$ \textbf{Enemy units}:\\
For example:\\
\{ \\
At point B1, there are: 16 SCV are gathering minerals in Orbital Command, 3 SCV are gathering gas in Refinery1, 3 SCV are gathering gas in Refinery2, 5 Marine are idling, 5 Marauder are idling, 3 Medivac are idling;

At point B2, there are: 18 SCV are gathering minerals in Command center, 3 SCV are gathering gas in Refinery1, 3 SCV are gathering gas in Refinery2;

At point A4, there are: 8 Marine are moving, 2 Siege Tank are moving, 4 Marauder are moving, 1 Cyclone are moving, 2 Reaper are moving, 1 Raven are moving, 2 viking are moving;

…\\
…\\
\} \\
$\bullet$ Enemy Buildings:\\
\{ \\
At point B1, there are: 1 Orbital Command, 2 Refinery, 3 Barracks, 2 Factory, 1 Engineering Bay, 9 Supply Depot, 2 Missile Turret, 1 Sensor Tower, 2 Starport, 1 Ghost Academy, 1 Armory;

At point B2, there are: 1 Command center, 1 Bunker, 2 Refinery, 2 Missile Turret;

At point B3, there are: Nothing;\\
…\\
…\\
\} \\
(6) Response rules.\\
(7) Response format.\\

\newpage

\subsection*{A.1.2 The full prompt of Overmind Brain}

\subsection*{A.1.2.1 The full prompt of Overmind Brain without critical battlefield information}

Note that the following variables expect for <pre\_thoughts> (comes from the previous round of Overmind Brain's inference) are retrieved through the python-sc2 interface and processed to natural language from the in-game state:\\
$\bullet$ <pre\_thoughts>\\
$\bullet$ <cur\_units>\\
$\bullet$ <cur\_buildings>\\
$\bullet$ <cur\_abilities>\\
$\bullet$ <enemy\_units>\\
$\bullet$ <enemy\_buildings>\\

\definecolor{lightgray}{gray}{0.9}
\begin{tcolorbox}[
  breakable, 
  colback=lightgray, 
  colframe=black, 
  left=10pt, 
  right=10pt, 
  top=10pt, 
  bottom=10pt, 
  boxrule=0.5pt, 
  width=\textwidth, 
  arc=0pt, 
  outer arc=0pt, 
  fontupper=\ttfamily 
]

You are an intelligent brain of Zerg swarm in StarCraft II game. You are very aggressive and know all the dependencies between Zerg units, Zerg buildings, and Zerg technological research.\\
You and Terran are on a square map with 16 mines evenly distributed on the map, as shown in the following matrix:\\

$\begin{array}{l}
[[0, A6, B7, B3, B2], \\
\space [A5, 0, B8, B4, B1], \\
\space [A1, A4, A8, 0, B5], \\
\space [A2, A3, A7, B6, 0]]
\end{array}$\\

Currently, your base is at position A1 and the Terran base is at position B1. 0 represents no mine, and other letters represent the mine number.\\

------Your commands in previous round\\
<pre\_thoughts>\\
------\\

------Rule\\
You need to analyze the game progression by following a structured protocol based on the game situation. Make sure to include the following perspectives in your analysis:\\
1. Current Stage of the Match: Determine the current game stage based on our situation and opponent’s status you detected, whether it’s early, middle, or late stage.\\
2. The Condition of Our Forces: Assess our current status in dimensions of:\\
2.1 Our Zerg Units and Buildings: Scrutinize the state of Zerg Units and Buildings.\\
2.2 Our Zerg Technology: Analyze the current status of the Zerg technological research based on the unlocked research. \\
3. Our current Zerg Operational Strategy: Devise a reasonable strategy based on our current situation, opponent’s situation, and scouting intel.\\
4. Opponent's situation: Assess opponent’s current status in dimensions of:\\
4.1 Opponent's Units and Buildings: Analyze state of opponent’s Units and Buildings.\\
4.1 Opponent's Tactical Plan and our Potential Risks: Based on detected opponent’s Units and Buildings, predict the opponent's attack timing to prepare defensive measures in advance.\\
5. Scouting Intel: Stress the importance of recent and consistent scouting reports to stay updated on the enemy's unit composition, positioning, and possible incoming attacks or expansions.\\
6. Repeated instructions between current decisions and commands from previous round: Eliminate the impact of resource waste caused by repeated instructions. Before issuing an instruction, first check the previous instruction list in "Your commands in previous round". Please confirm which instructions do not need to be executed repeatedly. For example, instructions such as building a Spawnling Pool do not need to be executed repeatedly, while morphing Zergling may need to be executed repeatedly, so that resources will not be wasted doing repetitive things.
    
------\\

------Current battle situation\\
    
Your current Units consists of:\\
\{\\
<cur\_units>\\
\}\\
    
Your current Buildings consists of: \\
\{\\
<cur\_buildings>\\
\}\\
    
Your Zerg has developed these technological research: \\
\{\\
<cur\_abilities>\\
\}\\
    
You have detected enemy units in:\\
\{\\
<enemy\_units>\\
\}\\
    
You have detected enemy buildings in:\\
\{\\
<enemy\_buildings>\\
\}\\
------\\
 
Based on the current battle situation and Units and Buildings from both sides, a brief step-by-step analysis can be done from our strategy, Units and Buildings, economic and technical perspectives. Then, formulate 20 actionable, specific decisions from the following action list. These decisions should be numbered from 0, denoting the order in which they ought to be executed, with 0 signifying the most immediate and crucial action. 
For instance:\\

RESPONSE FORMAT:\\
1. Current Stage of the Match;\\
2. Our Zerg Units and Buildings;\\
3. Our current Zerg Operational Strategy;\\
4. Opponent's Units and Buildings;\\
5. Opponent's Tactical Plan and our Potential Risks;\\
6. Scouting Intel;\\
7. Eliminate Repeated instructions;\\
8. Based on above analysis, each instruction should contain the target location, like A1, A2, etc.\\
\{\\
‘0’: <IMMEDIATE\_ACTION\_TO\_BE\_EXECUTED\_FIRST>,\\
‘1’: <NEXT\_ACTION\_TO\_BE\_EXECUTED>,\\
‘2’: <SUBSEQUENT\_ACTION\_TO\_BE\_EXECUTED>, \\
... \\
\}\\

\end{tcolorbox}

\newpage
\subsection*{A.1.2.2 The full prompt of Overmind Brain with critical battlefield information}
In pursuit of enhancing the strategic inference capabilities of the SwarmBrain when confronted with routine updates, the utilization of the CoT methodology has been employed. However, encounters with exigent scenarios where time is of the essence have revealed that the CoT approach may inadvertently extend the duration of reasoning, leading to potential detrimental impacts. To address this challenge and ensure a timely response to critical battlefield information, a secondary prompt has been introduced as an input for the Overmind Brain. While bypassing the explicit articulation of the CoT outcomes may somewhat diminish the overall effectiveness, this adaptation significantly reduces the reasoning timeframe by half. Consequently, this approach offers a viable solution to efficiently manage time-critical circumstances.\\
Note that the following variables expect for <pre\_thoughts> (comes from the previous round of Overmind Brain's inference) are retrieved through the python-sc2 interface and processed to natural language from the in-game state:\\
$\bullet$ <pre\_thoughts>\\
$\bullet$ <critical\_battlefield\_information>\\
$\bullet$ <cur\_units>\\
$\bullet$ <cur\_buildings>\\
$\bullet$ <cur\_abilities>\\
$\bullet$ <enemy\_units>\\
$\bullet$ <enemy\_buildings>\\

\definecolor{lightgray}{gray}{0.9}
\begin{tcolorbox}[
  breakable, 
  colback=lightgray, 
  colframe=black, 
  left=10pt, 
  right=10pt, 
  top=10pt, 
  bottom=10pt, 
  boxrule=0.5pt, 
  width=\textwidth, 
  arc=0pt, 
  outer arc=0pt, 
  fontupper=\ttfamily 
]

You are an intelligent brain of Zerg swarm in StarCraft II game. You are very aggressive and know all the dependencies between Zerg units, Zerg buildings, and Zerg technological research.\\
You and Terran are on a square map with 16 mines evenly distributed on the map, as shown in the following matrix:\\

$\begin{array}{l}
[[0, A6, B7, B3, B2], \\
\space [A5, 0, B8, B4, B1], \\
\space [A1, A4, A8, 0, B5], \\
\space [A2, A3, A7, B6, 0]]
\end{array}$\\

Currently, your base is at position A1 and the Terran base is at position B1. 0 represents no mine, and other letters represent the mine number.\\

------Your commands in previous round\\
<pre\_thoughts>\\
------\\

------Rule\\
You need to analyze the game progression by following a structured protocol based on the game situation. Make sure to include the following perspectives in your analysis:\\
1. Current Stage of the Match: Determine the current game stage based on our situation and opponent’s status you detected, whether it’s early, middle, or late stage.\\
2. The Condition of Our Forces: Assess our current status in dimensions of:\\
2.1 Our Zerg Units and Buildings: Scrutinize the state of Zerg Units and Buildings.\\
2.2 Our Zerg Technology: Analyze the current status of the Zerg technological research based on the unlocked research. \\
3. Our current Zerg Operational Strategy: Devise a reasonable strategy based on our current situation, opponent’s situation, and scouting intel.\\
4. Opponent's situation: Assess opponent’s current status in dimensions of:\\
4.1 Opponent's Units and Buildings: Analyze state of opponent’s Units and Buildings.\\
4.1 Opponent's Tactical Plan and our Potential Risks: Based on detected opponent’s Units and Buildings, predict the opponent's attack timing to prepare defensive measures in advance.\\
5. Scouting Intel: Stress the importance of recent and consistent scouting reports to stay updated on the enemy's unit composition, positioning, and possible incoming attacks or expansions.\\
6. Repeated instructions between current decisions and commands from previous round: Eliminate the impact of resource waste caused by repeated instructions. Before issuing an instruction, first check the previous instruction list in "Your commands in previous round". Please confirm which instructions do not need to be executed repeatedly. For example, instructions such as building a Spawnling Pool do not need to be executed repeatedly, while morphing Zergling may need to be executed repeatedly, so that resources will not be wasted doing repetitive things.

------\\

------Current battle situation\\

<critical\_battlefield\_information>\\
    
Your current Units consists of:\\
\{\\
<cur\_units>\\
\}\\
    
Your current Buildings consists of: \\
\{\\
<cur\_buildings>\\
\}\\
    
Your Zerg has developed these technological research: \\
\{\\
<cur\_abilities>\\
\}\\
    
You have detected enemy units in:\\
\{\\
<enemy\_units>\\
\}\\
    
You have detected enemy buildings in:\\
\{\\
<enemy\_buildings>\\
\}\\
------\\
 
Based on the current battle situation and Units and Buildings from both sides, a brief step-by-step analysis can be done from our strategy, Units and Buildings, economic and technical perspectives. Then, formulate many actionable, specific decisions from the following action list. These decisions should be numbered from 0, denoting the order in which they ought to be executed, with 0 signifying the most immediate and crucial action. 
For instance:\\

RESPONSE FORMAT:\\
1. Current Stage of the Match, but do not print it;\\
2. Our Zerg Units and Buildings, but do not print it;\\
3. Our current Zerg Operational Strategy, but do not print it;\\
4. Opponent's Units and Buildings, but do not print it;\\
5. Opponent's Tactical Plan and our Potential Risks, but do not print it;\\
6. Scouting Intel, but do not print it;\\
7. Eliminate Repeated instructions, but do not print it;\\
8. Based on above analysis, each instruction should contain the target location, like A1, A2, etc.\\
\{\\
‘0’: <IMMEDIATE\_ACTION\_TO\_BE\_EXECUTED\_FIRST>,\\
‘1’: <NEXT\_ACTION\_TO\_BE\_EXECUTED>,\\
‘2’: <SUBSEQUENT\_ACTION\_TO\_BE\_EXECUTED>, \\
... \\
\}\\

\end{tcolorbox}

\newpage
\subsection*{A.2 The SC2 Brain}
\subsection*{A.2.1 The components of the prompt for SC2 Brain}

The input prompt of gpt-3.5-turbo consists of several components:\\
(1) The task of the LLM.\\
(2) Your current thought, which is the output of the Overmind Brain.\\
(3) Response rules.\\
(4) Response format.\\

\subsection*{A.2.2 The full prompt of SC2 Brain}
The full prompt for the SC2 Brain is shown below.\\
Note that the following variable <cur\_thoughts> are retrieved from the output of the Overmind Brain.\\

\begin{tcolorbox}[
  breakable, 
  colback=lightgray, 
  colframe=black, 
  left=10pt, 
  right=10pt, 
  top=10pt, 
  bottom=10pt, 
  boxrule=0.5pt, 
  width=\textwidth, 
  arc=0pt, 
  outer arc=0pt, 
  fontupper=\ttfamily 
]

You are a professional Zerg player in StarCraft II. You know all the dependencies between Zerg units, Zerg buildings, and Zerg technological research. Now, you are playing Zerg against Computer Terran player.\\

Your current thoughts:\\
\{\\
<cur\_thoughts>\\
\}\\

Based on “Your current thoughts”, you need to specify more detailed execution plans. It needs to be listed in numerical order. You should only respond in the format as described below:\\

RESPONSE FROMAT:\\
(Command Target)->(Action)->(Target)\\

------example\\
makefileCopy json code\\
\{\\
‘0’: (Zergling, A1)->(Move)->(A4), //It means send Zerglings at A1 to A4\\
‘1’: (Drone, A1)->(Gather gas)->(Extractor1, A1) // It means send Drones at A1 to gathering gas at Extractor1 at A1\\
‘2’: (Zergling, A1)->(Morph)->(Baneling) //It means Zergling at A1 need to morph to Baneling\\
... \\
\}\\
------\\

\end{tcolorbox}

\newpage
\subsection*{A.3 Examples}
In this section, we present the inferential outcomes of SwarmBrain under a variety of scenarios.
\subsection*{A.3.1 Early stage of the match (ZvT)}
When the match started, for Zerg units, there are only 12 Drones mining at A1, and 1 Overlord idling at A1. For Zerg buildings, there is only a Hatchery at A1. And the status of the enemy is unknown.
The sample output of the Overmind Brain is shown as follows:

1. Current Stage of the Match:\\
- Based on the information provided, it is still early in the game as there are only 12 Drones mining in the Hatchery and limited information about the opponent's units and buildings.\\

2. The Condition of Our Forces:\\
2.1 Our Zerg Units and Buildings:\\
- At point A1, there is 1 Hatchery and 12 Drones mining.\\
- At point A2, there is nothing.\\
- At point A3, there is nothing.\\
- At point A4, there is nothing.\\

2.2 Our Zerg Technology:\\
- No technological research has been unlocked yet.\\

3. Our current Zerg Operational Strategy:\\
\textbf{Since it is still early in the game, the current strategy should focus on expanding the economy and building up the Zerg forces.} This can be achieved by training more Drones, expanding to additional bases, and researching Zerg technologies to unlock more powerful units.\\

4. Opponent's situation:\\
4.1 Opponent's Units and Buildings:\\
- The enemy units and buildings have not been detected yet.\\

4.2 Opponent's Tactical Plan and our Potential Risks:\\
Without information about the opponent's units and buildings, it is difficult to predict their tactical plan or potential risks. \textbf{However, it is important to be prepared for any potential attacks or expansions by scouting the enemy base and gathering information.}\\

5. Scouting Intel:\\
- \textbf{Emphasize the importance of consistent scouting to gather information about the enemy's unit composition, positioning, and potential threats.}\\
- \textbf{Send Overlords or other units to scout the opponent's base and expansions regularly.}

6. Repeated instructions between current decisions and commands from previous round:
- Since there were no commands issued in the previous round, there are no repeated instructions to eliminate.

7. Based on the above analysis, the following actionable decisions can be made:\\
\{\\
'0': Send an Overlord to scout the opponent's base B1,\\
'1': Build additional Drones in Hatchery A1,\\
'2': Build a Spawning Pool in Hatchery A1,\\
'3': Train Zerglings from the Spawning Pool A1,\\
'4': Research Metabolic Boost upgrade at the Spawning Pool A1,\\
'5': Expand to a nearby mineral field A2,\\
'6': Build an Extractor on the gas geyser A2,\\
'7': Train Queens from the Hatchery A1 for larva injects and defense,\\
'8': Build a Roach Warren in Hatchery A1,\\
'9': Train Roaches from the Roach Warren A1,\\
'10':\textbf{ Upgrade Roach speed at the Roach Warren A1},\\
'11': Build an Evolution Chamber in Hatchery A1,\\
'12': Research Missile Attacks upgrade at the Evolution Chamber A1,\\
'13': Build a Lair in Hatchery A1 for advanced tech options,\\
'14': \textbf{Scout the opponent's expansions and map control with Zerglings},\\
'15': Build additional Overlords to maintain supply,\\
'16': Expand to another mineral field A3 for increased income,\\
'17': Build a Spire in Lair A1 for air unit production,\\
'18': Research Flyer Attacks upgrade at the Spire A1,\\
'19': Train Mutalisks from the Spire A1 for harassment and map control.\\
\}\\

In the initial analysis of the inferential outcomes derived from the Overmind Brain, it is discernible that at the inception of gameplay, it deduces a strategic emphasis on expanding the economy and building up the Zerg forces. Concurrently, considering the unknow status of the adversary's position and capabilities, the Overmind Brain exhibits a predilection for deploying Overlords to scout the enemy's initial localization. Notwithstanding this tactic, it is worth noting that it is atypical among professional players to dispatch an Overlord to the opponent's primary mineral field at the game's outset. This is chiefly due to the Overlord's intrinsic sluggish mobility during early play, which significantly increases the likelihood of its rapid neutralization by hostile forces. Nevertheless, this strategy illuminates the Overmind Brain's acute awareness of the pivotal role intelligence gathering plays in strategic gameplay.

The json action list is used as the <cur\_thoughts> to the SC2 Brain, the example output of the SC2 Brain is shown as follows:

\{\\
'0': "(Overlord, A1)->(Move)->(B1)",\\
'1': "(Drone, A1)->(Train)->(Drone)",\\
'2': "(Drone, A1)->(Build)->(Spawning Pool, A1)",\\
'3': "(Spawning Pool, A1)->(Train)->(Zergling)",\\
'4': "(Spawning Pool, A1)->(Research)->(Metabolic Boost)",\\
'5': "(Drone, A1)->(Build)->(Hatchery, A2)",\\
'6': \textbf{"(Drone, A2)->(Build)->(Extractor, A2)"},\\
'7': "(Hatchery, A1)->(Train)->(Queen)",\\
'8': "(Drone, A1)->(Build)->(Roach Warren)",\\
'9': "(Roach Warren, A1)->(Train)->(Roach)",\\
'10': \textbf{"(Roach Warren, A1)->(Research)->(Roach Speed)"},\\
'11': "(Drone, A1)->(Build)->(Evolution Chamber)",\\
'12': "(Evolution Chamber, A1)->(Research)->(Missile Attacks)",\\
'13': "(Hatchery, A1)->(Build)->(Lair)",\\
'14': \textbf{"(Zergling, A1)->(Move)->(Opponent's expansions and map control)"},\\
'15': "(Hatchery, A1)->(Train)->(Overlord)",\\
'16': "(Drone, A1)->(Build)->(Hatchery, A3)",\\
'17': "(Lair, A1)->(Build)->(Spire, A1)",\\
'18': "(Spire, A1)->(Research)->(Flyer Attacks)",\\
'19': "(Spire, A1)->(Train)->(Mutalisk)"\\
\}

Since the Overmind Brain and SC2 Brain relay on the LLM, it occasionally exists small problems. 
For instance, the tenth command "Upgrade Roach speed at the Roach Warren A1" generated by the Overmind Brain reveals a misunderstanding of the Roach Warren's capabilities, as it does not provide an upgrade of Roach velocity. Consequently, such a command remains unexecutable and exerts minimal impact on the overarching strategy.
Complications also arise with the fourteenth command, "Scout the opponent's expansions and map control with Zerglings." Here, the Overmind Brain does not specify the exact coordinates of the opponent, leading to an ambiguous interpretation of SC2 Brain wherein the coordinates are described simply as "Opponent's expansions and map control." To address this ambiguity, a methodological approach has been adopted—where, in the absence of concrete coordinates, a search is performed using synonyms or key terms akin to 'opponent'. Once a match is discerned, the directive is then assigned the coordinates B1 to maintain operational clarity.
Small issues also extend to SC2 Brain's inferences. For example, the sixth command entails dispatching a Drone from location A2 to construct an Extractor. However, there currently exists no Drone at A2. To rectify such discrepancies, our method ensures that when no Drone from A2 can be located, an available Drone from another location is reallocated to fulfill the construction task. This response serves to bridge the gap between planning and execution within the game's operational logic.

\newpage
\subsection*{A.3.2 When detected a group of enemy (ZvT)}

In the early stages of gameplay, following a period of development, SwarmBrain has reached the following state: \\
Zerg unit state:\\
At point A1, there are: 10 Drones are gathering minerals in Hive, 3 Drone are gathering gas in Extractor1, 3 Drone are gathering gas in Extractor2, 5 Overlord are idling, 1 Queen constantly injecting eggs into Hive;\\
At point A2, there are: 19 Drones are gathering minerals in Hatchery, 3 Drone are gathering gas in Extractor1, 3 Drone are gathering gas in Extractor2, 4 Overlord are idling, 1 Queen constantly injecting eggs into Hatchery;\\
At point A3, there are: 1 Overlord are idling;\\
At point A4, there are: 10 Drones are gathering minerals in Hatchery, 3 Drone are gathering gas in Extractor1, 3 Drone are gathering gas in Extractor2, 1 Queen constantly injecting eggs into Hatchery, 11 Hydralisk are idling, 13 Roach are idling, 14 Zergling are idling;\\
At point A5, there are: 1 Overlord are idling;\\
At point A7, there are: 1 Overlord are idling;\\
Zerg building state:\\
At point A1, there are: 1 Hive, 2 Extractor, 1 Infestation Pit, 1 Evolution Chamber, 1 Roach Warren, 1 Hydralisk Den, 1 Spawning Pool, 1 Spire, 1 Nydus Network, 1 Spine Crawler, 1 Spore Crawler;\\
At point A2, there are: 1 Hatchery, 2 Extractor; \\
At point A3, there are: Nothing;\\
At point A4, there are: Nothing;\\

Upon the Overlord's detection of enemy proximity at A4, the Overmind Brain articulates the following inference:

\{\\
'0': \textbf{"Reinforce units at A4 with additional Zerglings and Roaches"},\\
'1': "Research additional Zerg upgrades at the Evolution Chamber",\\
'2': \textbf{"Train additional Zerglings and Roaches at A4 to reinforce the defense"},\\
'3': "Build more Spine Crawlers and Spore Crawlers at A4 for additional defense",\\
'4': \textbf{"Train more Hydralisks at A4 to counter the Terran air units"},\\
'5': "Coordinate with nearby Overlords to provide vision and detection around A4",\\
'6': \textbf{"If the defense is successful, launch a counter-attack on the enemy base at B1 or B2"}\\
\}

The inferential analysis reveals that upon detection of an emergent adversarial threat at A4, Overmind Brain promptly responds by reinforcing the offensive capabilities at A4. This reinforcement strategy involves mobilization for augmenting the combat forces, which includes rallying additional Zerglings, Roaches, and Hydralisks, as well as initiating upgrades and research—despite the fact that it may be too late, and some research has already been completed in prior inferences. Overmind Brain also issues orders to deploy Overlords for reconnaissance. Furthermore, it issues commands for a counteroffensive targeting the enemy's B1 and B2 sectors upon successful defense. This strategy is deemed judicious in the context of swiftly concluding hostilities, as a well-timed counterattack following the depletion of an enemy's offensive power can be instrumental in expediently bringing the conflict to an end. 

The json action list is used as the <cur\_thoughts> to the SC2 Brain, the example output of the SC2 Brain is shown as follows:

\{\\
'0': "(Zerglings, A4)->(Reinforce)->(A4)",\\
'1': "(Roaches, A4)->(Reinforce)->(A4)",\\
'2': "(Evolution Chamber, A4)->(Research)->(Zerg upgrades)",\\
'3': "(Larva, A4)->(Train)->(Zerglings)",\\
'4': "(Larva, A4)->(Train)->(Roaches)",\\
'5': "(Spine Crawlers, A4)->(Build)->(A4)",\\
'6': "(Spore Crawlers, A4)->(Build)->(A4)",\\
'7': "(Larva, A4)->(Train)->(Hydralisks)",\\
'8': "(Overlords, A4)->(Coordinate)->(A4)",\\
'9': "(Overlords, A4)->(Coordinate)->(B1)",\\
'10': "(Overlords, A4)->(Coordinate)->(B2)",\\
'11': "(Zerg units, A4)->(Launch counter-attack)->(Enemy base at B1)",\\
'12': "(Zerg units, A4)->(Launch counter-attack)->(Enemy base at B2)"\\
\}

\end{document}